\documentclass[lettersize,journal]{IEEEtran}
\IEEEpubidadjcol
\usepackage{amsmath,amsfonts}
\usepackage{algorithmic}
\usepackage{algorithm}
\usepackage{array}
\usepackage[caption=false,font=normalsize,labelfont=sf,textfont=sf]{subfig}
\usepackage{textcomp}
\usepackage{stfloats}
\usepackage{url}
\usepackage{verbatim}
\usepackage{graphicx}
\usepackage{cite}
\usepackage{color}
\usepackage{bbm}
\usepackage{multirow}
\usepackage[table,xcdraw]{xcolor}
\usepackage{booktabs}
\usepackage{bbding}
\usepackage{makecell}
\usepackage{hyperref} 
\ifCLASSOPTIONcompsoc
\usepackage[caption=false, font=footnotesize, labelfont=sf, textfont=sf]{caption}
\else
\usepackage[font=footnotesize]{caption}
\fi
\newcommand{\rf}[1]{{\textbf{\color{red}{#1}}}} 
\newcommand{\bd}[1]{{\color{blue}{\underline{#1}}}} 
\newcommand{\insertfig}{\setcounter{figure}{0}
\includegraphics[width=\linewidth]{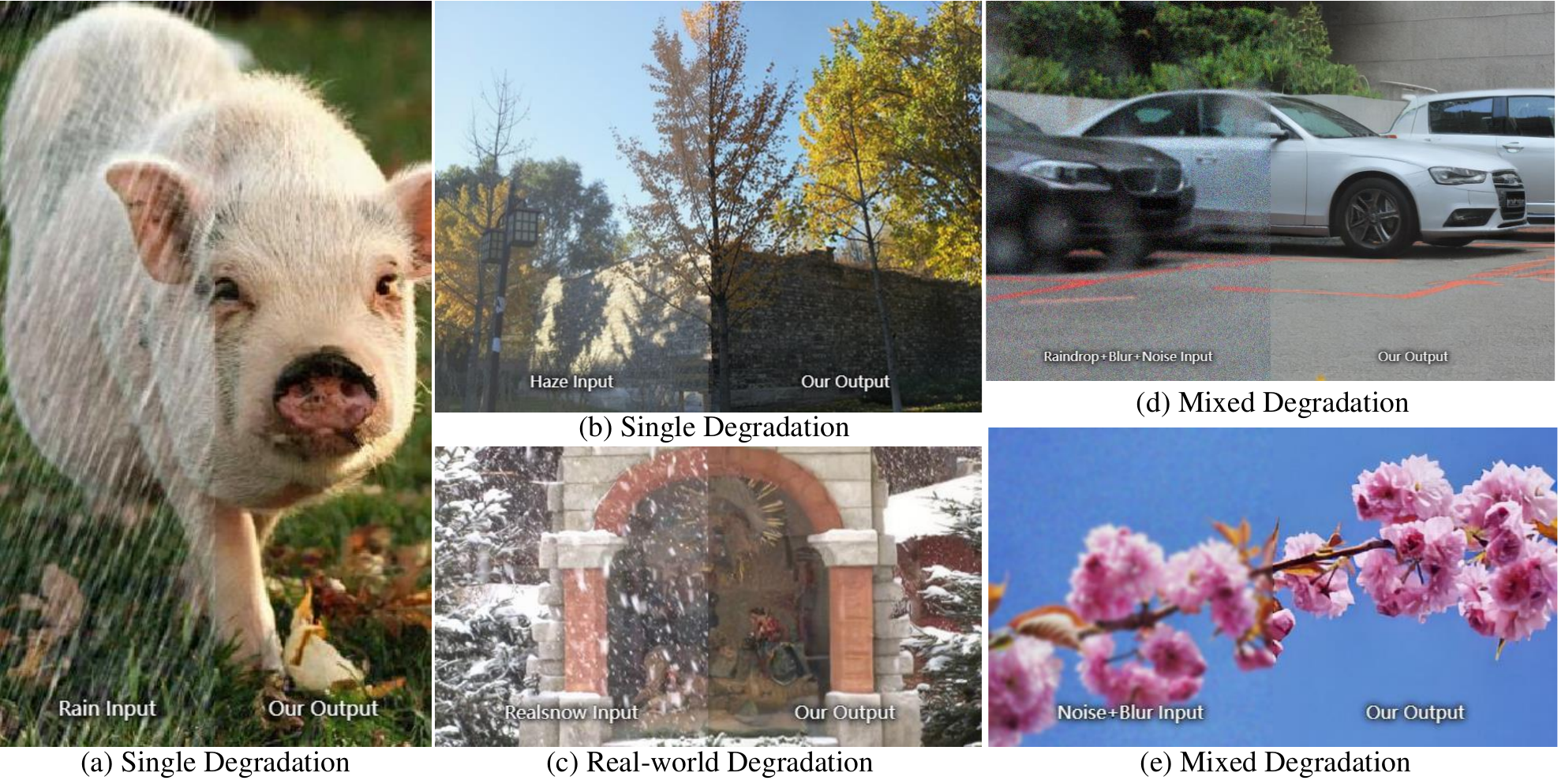} \vspace{-0.6cm}
\captionof{figure}{
Our Diff-Restorer model demonstrates remarkable restoration results on the restoration tasks with single degradation, real-world degradation, and mixed degradation. Diff-Restorer can adaptively handle various degradation types.}
   \label{fig:teaser}\vspace{-0.9cm}}

\makeatletter
\apptocmd{\@maketitle}{\centering\insertfig}{}{}
\makeatother
\begin{document}

\title{Diff-Restorer: Unleashing Visual Prompts for Diffusion-based Universal Image Restoration}

\author{Yuhong Zhang, Hengsheng Zhang, Xinning Chai, Zhengxue Cheng, Rong Xie,~\IEEEmembership{Member,~IEEE}, Li Song,~\IEEEmembership{Senior Member,~IEEE}, Wenjun Zhang,~\IEEEmembership{Fellow Member,~IEEE}
\thanks{Yuhong Zhang, Hengsheng Zhang, Xinning Chai, Zhengxue Cheng, Rong Xie and Wenjun Zhang are with
Institute of Image Communication and Network Engineering, Shanghai Jiao Tong University, China (e-mail: rainbowow@sjtu.edu.cn; hs\_zhang@sjtu.edu.cn; chaixinning@sjtu.edu.cn; zxcheng@sjtu.edu.cn, xierong@sjtu.edu.cn; zhangwenjun@sjtu.edu.cn).}
\thanks{Li Song is with Institute of Image Communication and Network Engineering, Shanghai Jiao Tong University and the MoE Key Lab of
Artificial Intelligence, AI Institute, Shanghai Jiao Tong University, China (email: song\_li@sjtu.edu.cn).}
}

\markboth{Journal of \LaTeX\ Class Files,~Vol.~14, No.~8, August~2021}%
{Shell \MakeLowercase{\textit{et al.}}: A Sample Article Using IEEEtran.cls for IEEE Journals}

\maketitle

\begin{abstract}
Image restoration is a classic low-level problem aimed at recovering high-quality images from low-quality images with various degradations such as blur, noise, rain, haze, etc. However, due to the inherent complexity and non-uniqueness of degradation in real-world images, it is challenging for a model trained for single tasks to handle real-world restoration problems effectively. Moreover, existing methods often suffer from over-smoothing and lack of realism in the restored results. To address these issues, we propose Diff-Restorer, a universal image restoration method based on the diffusion model, aiming to leverage the prior knowledge of Stable Diffusion to remove degradation while generating high perceptual quality restoration results. Specifically, we utilize the pre-trained visual language model to extract visual prompts from degraded images, including semantic and degradation embeddings. The semantic embeddings serve as content prompts to guide the diffusion model for generation. In contrast, the degradation embeddings modulate the Image-guided Control Module to generate spatial priors for controlling the spatial structure of the diffusion process, ensuring faithfulness to the original image. Additionally, we design a Degradation-aware Decoder to perform structural correction and convert the latent code to the pixel domain. We conducted comprehensive qualitative and quantitative analysis on restoration tasks with different degradations, demonstrating the effectiveness and superiority of our approach.
\end{abstract}

\begin{IEEEkeywords}
Image restoration, universal image restoration, diffusion models, visual prompt, CLIP.
\end{IEEEkeywords}

\section{Introduction}
\IEEEPARstart{I}{mage} restoration aims to recover clear and high-quality (HQ) images from degraded low-quality (LQ) images. Due to the diversity of degradation types, it involves various specific sub-tasks, including deblurring\cite{kupyn2018deblurgan, kupyn2019deblurgan}, denoising\cite{zhang2017denoise, ding2024tcsvtdenoisewavelet}, deraining\cite{liang2022derain, jiang202derain}, dehazing\cite{song2023dehaze, wang2024ucldehaze}, image enhancement\cite{zhang2023enhance, wang2024zero}, etc. Considering the diversity of these sub-tasks, previous methods \cite{kupyn2018deblurgan, zhang2017denoise, jiang202derain, wang2024ucldehaze, zhang2023enhance, wang2023stablesr} often trained individual task-specific models to tackle separate restoration problems. While these single-task methods have achieved good results within their respective task scopes, they face many challenges in practical applications due to the complexity and diversity of the real world. First, degradation in the real world is often more complex and does not perfectly match the predefined degradation types in the training data. Second, real-world images may not contain a single degradation but rather combinations of multiple degradations. Therefore, it is difficult to address them using methods designed for single-task scenarios. To tackle these problems, methods\cite{chen2022nafnet, zamir2022restormer} attempt to design a base model to handle different sub-tasks, but these methods require separate training for each specific task. There are also some models \cite{li2022airnet, valanarasu2022transweather} that attempt to solve the multi-task restoration problem within the same model framework, but they often consider only 3-4 restoration tasks and the restoration results are generally average. Methods like \cite{potlapalli2024promptir, li2023pip} attempt to address the multi-task restoration problem by learning implicit prompts, but due to limitations of the network itself, they exhibit subpar performance in generating details. Additionally, some methods\cite{conde2024instructir, liu2024diff-plugin} utilize instructions for image restoration, but human-based instruction methods face significant challenges as users must manually identify the degradation types present in the image and provide accurate instructions. However, for complex degradation images, users often struggle to provide correct instructions. Therefore, we hope to design an image restoration method that is applicable to various degradation tasks and is capable of adaptively extracting information from low-quality images for restoration.

What's more, with the development of image restoration techniques, there is an increasing expectation for the perceived quality of restored images, moving beyond simple degradation removal towards more realistic image restoration. Therefore, generative models such as Variational Autoencoders (VAEs) \cite{kingma2013vae}, Generative Adversarial Networks (GANs)\cite{goodfellow2014gan}, and Diffusion Models \cite{ho2020ddpm, rombach2022ldm} can be utilized to generate more realistic and natural images. However, these methods often require defining or learning an accurate explicit degradation function for each degradation type, which limits their widespread applicability in diverse and complex degradation scenarios. In the past two years, pre-trained Text-to-Image (T2I) models, such as Stable Diffusion (SD) \cite{rombach2022ldm}, have achieved unprecedented success in image generation and have been widely applied in downstream tasks such as image classification\cite{li2023sdclassifier}, segmentation\cite{karazija2023sdseg}, and editing\cite{brooks2023instructpix2pix}. SD is trained on a dataset of over 5 billion image-text pairs, providing rich natural image priors and serving as a vast texture and structure library. Therefore, a natural idea is to leverage SD as prior knowledge for image restoration. Some methods\cite{wang2023stablesr, lin2023diffbir, wu2023seesr} have attempted to use SD for blind image super-resolution. Mperceiver\cite{ai2023mperceiver} designed a dual-branch structure for multi-task image restoration. However, due to the inherent uncontrollability of SD, the aforementioned methods still need improvement in terms of generated quality, and ensuring consistency with the spatial information of the input image poses a significant challenge.

Based on the aforementioned challenges, we propose a diffusion-based universal image restoration method called \textbf{Diff-Restorer}. Visual language models \cite{radford2021clip, li2023blip} possess strong image perception and representation capabilities, while T2I diffusion models\cite{rombach2022ldm, saharia2022imagen} have significant advantages in generating high-quality images. Therefore, we combine the strengths of these two models to address low-level problems. Specifically, we utilize the visual prompts extracted from the visual language model CLIP\cite{radford2021clip}, which primarily include semantic information and degradation information. We use the semantic information as content prompts for the T2I diffusion model. Since restoration tasks differ from editing tasks, faithful preservation of the original image's spatial structure is required. To achieve this, we design a Image-guided Control Module with degradation modulation to provide spatial structure and color priors. The aforementioned approach is capable of generating detailed and high-quality restored images. However, due to the diffusion process of SD occurring in the latent domain, there may inevitably be some deformations in certain details. Therefore, we designed a Degradation-aware Decoder to perform corrections. To evaluate the effectiveness and universality of the proposed method, we conducted experiments on single degradation restoration tasks, real-world degradation restoration tasks, and mixed degradation restoration tasks. We have taken into account various common degradation types that occur in daily life, including noise, blur, low-light, haze, rain, raindrop, JPEG compression artifact, and snow. Through extensive evaluations, our Diff-Restorer achieved convincing results in terms of generality, adaptability, faithfulness, and perceptual quality. Fig. \ref{fig:teaser} illustrates the effects of our model, showcasing its superior performance.

Our contributions can be summarized as follows:
\begin{itemize}
\item{We propose Diff-Restorer, a method that adaptively extracts visual prompts from the input image and guides SD for different restoration tasks.}
\item{We utilize CLIP to extract semantic information and degradation information, using semantic information as content prompts and degradation information to modulate the Image-guided Control Module. Additionally, to address the issue of SD losing structural information in the latent domain, we design a Degradation-aware Decoder that performs structural correction.}
\item{We validate our proposed method through comprehensive experiments on single degradation restoration tasks, real-world degradation restoration tasks, and mixed degradation restoration tasks. Both qualitative and quantitative results demonstrate the effectiveness and generality of our Diff-Restorer.}
\end{itemize}

\section{Related Work}
\subsection{Universal Image Restoration}
The concept of universal image restoration has recently gained significant development. This field aims to utilize a unified model with a single set of pre-trained weights to handle various types of degradation. PromptIR~\cite{potlapalli2024promptir} and ProRes~\cite{ma2023prores} introduce additional degradation context to incorporate task-specific information and learn task-adaptive prompts to guide the network for multi-task restoration. PIP\cite{li2023pip} extends the notion of prompts by introducing degradation-aware prompts and base restoration prompts, enhancing existing image restoration models for tasks such as denoising, deraining, dehazing, deblurring, and low-light enhancement. AirNet~\cite{li2022airnet} and DASR~\cite{wang2021dasr} employ contrastive learning methods to design network constraints that help the network differentiate input images from different tasks and process them accordingly. The aforementioned works primarily focus on task adaptability. However, some methods consider alternative perspectives. IDR~\cite{zhang2023idr} explores model optimization through component-oriented clustering, investigating correlations between various restoration tasks using a component-based paradigm. TransWeather~\cite{valanarasu2022transweather} designs a transformer-based learnable weather type query network to handle different weather conditions. Zhu \emph{et al.}~\cite{zhu2023learning} propose a strategy that simultaneously considers overall weather characteristics and specific weather features. Yang \emph{et al.}~\cite{yang2023language} utilize a pre-trained visual language model (PVL) to discriminate degradation and assist in image restoration. Nevertheless, most existing methods often consider only a few types of degradation modeling, making it challenging to generalize them to real-world applications. The limited number of tasks fails to reflect training conflicts between different restoration tasks and can only handle a limited range of degradation types, thus failing to cover complex real-world scenarios. Our approach utilizes visual prompts from CLIP to adaptively address different restoration tasks.

\subsection{Diffusion-based Image Restoration}
Recently, Text-to-Image Diffusion Models, such as Stable Diffusion \cite{rombach2022ldm}, have achieved success in high-quality and diverse image synthesis. Due to the strong generative capabilities of diffusion models in producing realistic images~\cite{dhariwal2021dbg,ho2020ddpm,saharia2022imagen,ramesh2022t2i}, some diffusion-based methods have been proposed for image restoration. These methods can be primarily categorized into zero-shot and supervised learning-based approaches~\cite{li2023restorationsurvey}. Zero-shot methods, such as~\cite{chung2022improving,fei2023gdp}, utilize pre-trained diffusion models as generative priors and seamlessly incorporate degraded images as conditioning during the sampling process. Supervised learning-based methods, such as~\cite{li2022srdiff, ozdenizci2023weatherdiff}, train a conditional diffusion model from scratch. Recently, some methods~\cite{wang2023stablesr,yang2023pasd, lin2023diffbir, wu2023seesr} have attempted to use pre-trained T2I diffusion models for blind image super-resolution. Some works also attempt to utilize the T2I models for the universal image restoration task. TextualDegRemoval~\cite{lin2023textualdegremoval} approaches image restoration from a text restoration perspective, while AutoDIR~\cite{jiang2023autodir} proposes an approach that first evaluates image quality, generates restoration instructions, and then utilizes the diffusion model for restoration. DA-CLIP~\cite{luo2023da-clip} improves CLIP to generate degraded image type texts and high-quality semantic texts and introduces these texts into the diffusion model to guide multi-task image restoration. MPerceiver~\cite{ai2023mperceiver} introduces a dual-branch structure to leverage the diffusion model for all-in-one image restoration. Diff-Plugin~\cite{liu2024diff-plugin} has trained multiple additional components for different restoration tasks and utilizes language instructions to select the appropriate components for addressing different tasks. Although the aforementioned methods have achieved some success, there are still limitations in terms of image restoration quality and model generalization. Therefore, it motivates us to further explore the potential of SD for universal image restoration, leveraging its high-quality image priors to improve reconstruction quality and generalization in real-world scenarios and generate more high-quality realistic image details. Additionally, manual instructions may suffer from inaccuracies and other issues. Hence, we aim to fully exploit the information present in low-quality images and adaptively handle different restoration tasks instead of relying on manual instructions. Moreover, as a diffusion model operating in the latent space, Stable Diffusion adopts a highly compressed VAE architecture, which carries the risk of losing fine-grained details.

\begin{figure}[t] 
\centering
\includegraphics[width=0.48\textwidth]{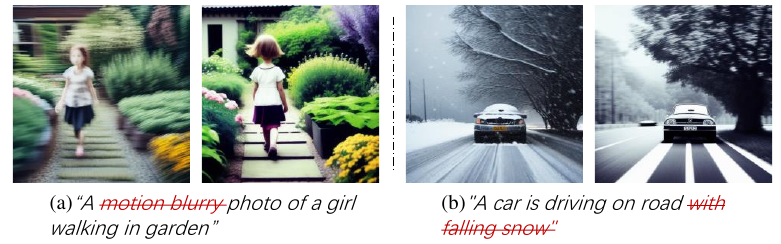}
\vspace{-2mm}
\caption{Examples of the generated images with SD img2img. The left image of each subfigure is generated by the full prompt. The right image is translated from the left image under the guidance of the prompt removing the degradation-related description highlighted in \textcolor[RGB]{192,0,0}{red}. (a) is an example of deblurring correlation, while (b) is an example of desnowing. SD img2img can achieve degradation removal but result in changes to the image content.}
\vspace{-3mm}
\label{fig: motivation_s}
\end{figure}

\section{Method}
\subsection{Motivation and Overview}
\textbf{Motivation.}
The primary goal of image restoration is to remove degradation and generate visually clear and realistic images. With the advancements in generative models, pre-trained T2I models, such as Stable Diffusion\cite{rombach2022ldm}, have demonstrated significant potential and advantages in content generation. This inspires us to utilize generative models for image restoration and generate more realistic images. Prompts are crucial factors influencing the quality of Stable Diffusion. We have investigated the impact of different prompts on the quality of image generation. As shown in Fig. \ref{fig: motivation_s}, we first generated the left images using the full prompts including the content descriptions and degradation-related descriptions highlighted in red. Then, we removed the degradation-related descriptions and used the SD img2img function\footnote{https://stablediffusion.fr/webui} to generate the right image, conditioning the left image generated by the full prompt. It can be observed that removing the degradation-related prompt also removes the degradation in the generated image. This inspires us to approach image restoration from a textual restoration perspective. However, this approach faces several challenges. On the one hand, it is difficult to generate a prompt that can fully reconstruct the original LQ image. Utilizing the existing caption generation models\cite{li2023blip} to generate a prompt does not completely capture the structure and content of the original image. Thus, simply removing the degradation description from this prompt for image generation results in significant differences from the original LQ image, which defeats the goal of image restoration. On the other hand, directly using the SD img2img function can lead to changes in the content, also deviating from the faithful representation of the original image. 

For restoration tasks, we expect the model to remove degradation while preserving other content. Therefore, we adopt an implicit approach to extract semantically relevant information from the image as SD's guidance and add control module to ensure fidelity to the input. Additionally, we aim for the model to adaptively recognize the degradation type of the input and utilize the identified degradation-related information to address potential training conflicts among different restoration tasks. 

\begin{figure}[t] 
\centering
\includegraphics[width=0.48\textwidth]{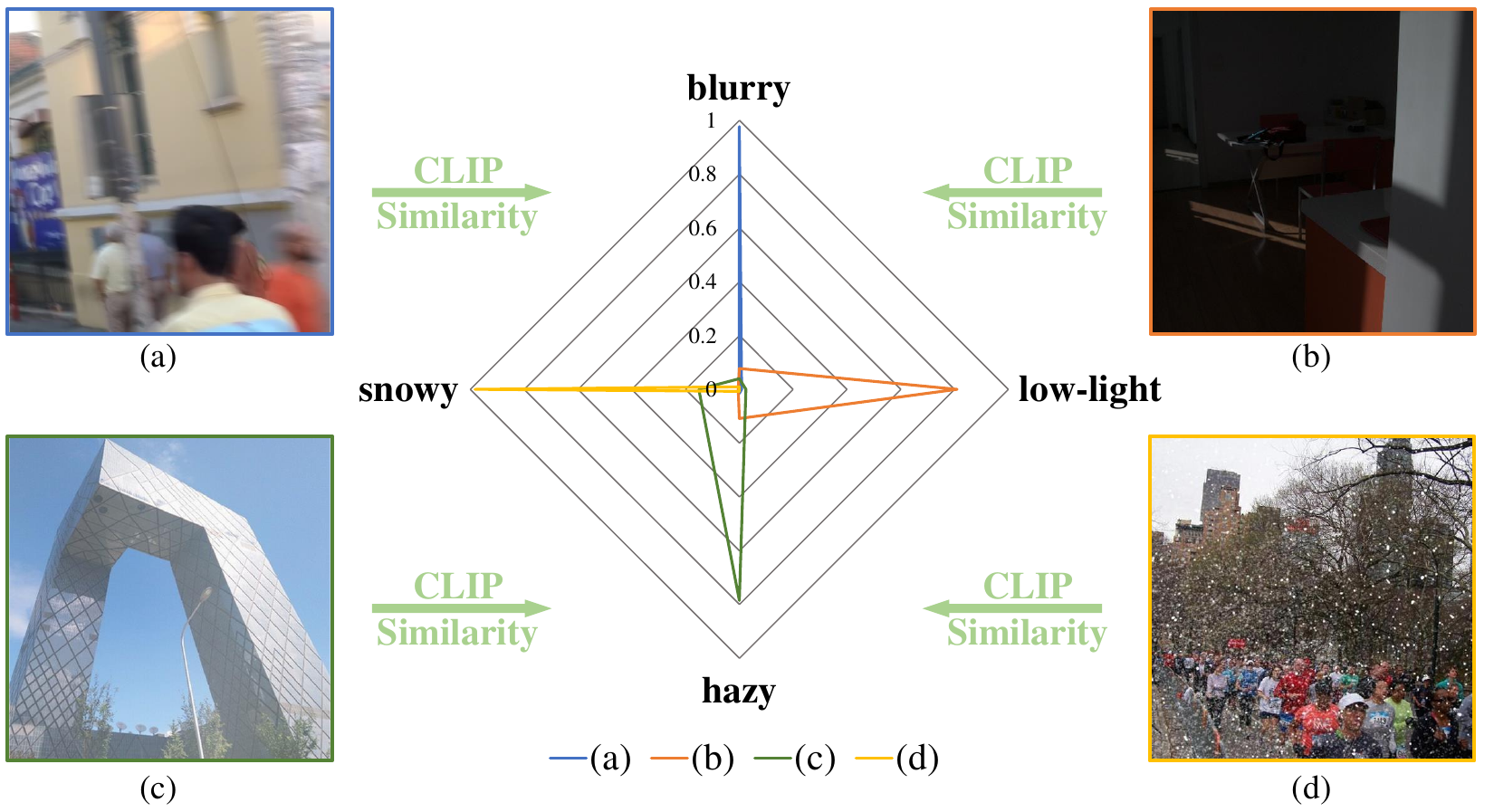}
\vspace{-2mm}
\caption{Illustration of the similarity between degraded images and degradation-related texts. (a)-(d) represent four different degradation types, and their similarity with four texts (\textit{``blurry photo"}, \textit{``low-light photo"}, \textit{``hazy photo"}, and \textit{``snowy photo"}) is shown in the middle.}
\vspace{-3mm}
\label{fig: motivation_d}
\end{figure}

Fortunately, CLIP\cite{radford2021clip}, as a pre-trained multimodal model learned from a large corpus of image-text pairs, possesses powerful visual representation capabilities and intrinsically encapsulates information about human perception. Therefore, it can be employed to evaluate the degradation type of an image. To validate our idea, we conducted a simple preliminary experiment to test whether CLIP can correctly assess the degradation type. Specifically, we randomly selected several images with different degradation types and defined prompts related to the degradation, such as \textit{``blurry photo"}, \textit{``low-light photo"}, etc. Following CLIP-IQA\cite{wang2023clipiqa}, we calculated the text-image similarity between the images and the textual prompts. The similarity is calculated as this:
\begin{equation}
s_i = \frac{e^{cos(\mathcal{E}_{clip}(I_{LQ}), \mathcal{E}_{clip}(T_i))}}{e^{cos(\mathcal{E}_{clip}(I_{LQ}), \mathcal{E}_{clip}(T_1))} + \dots+e^{cos(\mathcal{E}_{clip}(I_{LQ}), \mathcal{E}_{clip}(T_n))}},
\label{eq1}
\end{equation}
where $I_{LQ}$ denotes the input image, $T_i$ denotes the $i_{th}$ text, $n$ is the number of the texts, $s_i$ denotes the similarity between the $I_{LQ}$ and $T_i$, $\mathcal{E}_{clip}$ denotes the CLIP encoder, and $cos(\cdot, \cdot)$ denotes the cosine similarity calculation.

As shown in Fig. \ref{fig: motivation_d}, the similarity between the images and degradation-related prompts is consistent with human perception. The highest similarity significantly surpasses the similarity to other degradation descriptions. This experimental observation indicates that CLIP can serve as an indicator to distinguish different degradation types of images. 

\begin{figure*}[t]
\centering
\includegraphics[width=0.95\textwidth]{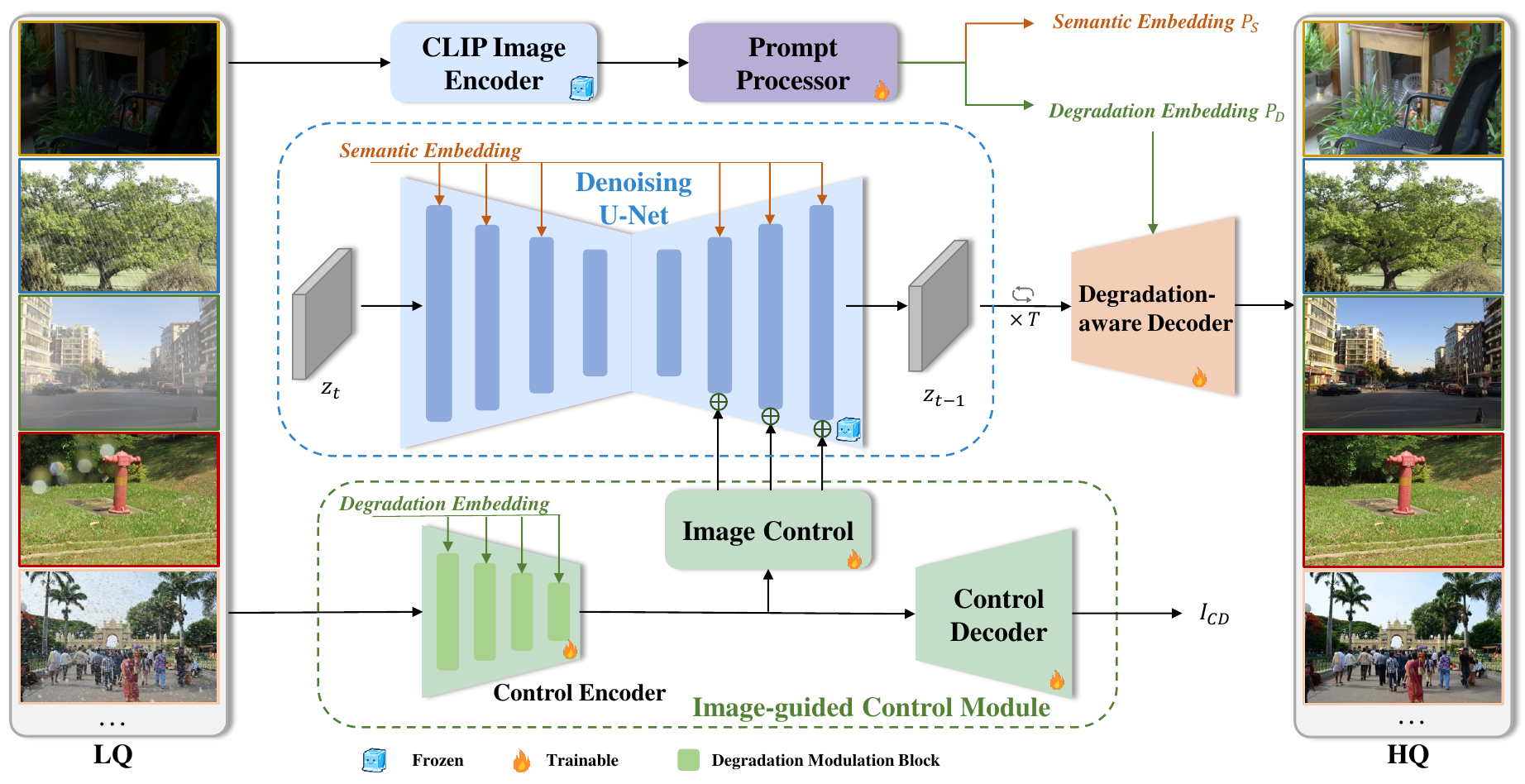}
\vspace{-2mm}
\caption{Overview of the architecture of Diff-Restorer. CLIP Image Encoder and Prompt Processor are used to extract visual prompts, which include semantic embedding and degradation embedding. A Image-guided Control Module modulated with degradation embedding is used to provide control information. The pre-trained denoising U-Net utilizes control information and semantic embedding as conditions to denoise. After multiple denoising steps, the latent features are generated and subsequently transformed into high-quality restored images by the Degradation-aware Decoder.}
\label{fig: overview}
\vspace{-3mm}
\end{figure*}

\textbf{Overview.}
Based on the above observation, we propose \textbf{Diff-Restorer}, a method aiming to extract semantic and degradation information from degraded images and utilize the pre-trained T2I models to generate high-quality and realistic restored images. The overall framework of the proposed Diff-Restorer is illustrated in Fig. \ref{fig: overview}. First, we utilize CLIP to extract image embeddings and design a \textbf{Prompt Processor} to extract visual prompts, including semantic embedding and degradation embedding. The semantic embedding is fed into the denoising U-Net as guidance for image generation and the degradation embedding serves as a identifier to distinguish different tasks. Additionally, to ensure that the generated images faithfully preserve the spatial structure of the original images, we introduce a \textbf{Image-guided Control Module} to exploit high-quality information from low-quality images and control the spatial structure and color of the generated images. To adaptively handle different degradation types and address potential training conflicts in different restoration tasks, we employ the degradation embedding to modulate the control module. Finally, since the diffusion process of Stable Diffusion occurs in the latent domain, certain spatial information is inevitably lost. To address this, we separately train a \textbf{Degradation-aware Decoder} to refine the generated images and produce higher-quality restoration results. The aforementioned modules cooperate with each other to assist the diffusion model in image restoration.

\subsection{Visual Prompt Processor}
CLIP possesses powerful representational capabilities for images and contains rich and meaningful information about images. Therefore, we utilize the pre-trained CLIP image encoder to extract visual information which is used to aid in image restoration. For the universal image restoration task, we aim to extract both semantic information and degradation information from the low-quality image. The semantic information is fed into the U-Net of the diffusion model through cross-attention, guiding the image generation process. The degradation information helps to differentiate between different degradation tasks and assist the Image-guided Control Module for the restoration task. For semantic information, we expect it to solely capture the content of the input image without including any degradation information. If the input image $I_{LQ}$ is degraded, the image embedding $P_{CLIP}$ from CLIP image encoder $\mathcal{E}_{clip}$ will inherently contain corresponding degradation information. Due to the presence of degradation in the embedding, the synthesized image will inevitably reflect the associated degradation patterns. Therefore, we propose a Prompt Processor that consists of two branches: a semantic branch $\mathcal{B}_s$ for extracting semantic information $P_{S}$ and a degradation branch $\mathcal{B}_d$ for identifying degradation information $P_{D}$. The overall process is summarized as:
\begin{equation}
\begin{aligned}
    P_{CLIP} &= \mathcal{E}_{clip}(I_{LQ}), \\ 
   P_{S} &= \mathcal{B}_s(P_{CLIP}), \\ 
   P_{D} &= \mathcal{B}_d(P_{CLIP}), 
\end{aligned}
\label{eq2}
\end{equation}
where semantic embedding $P_S \in \mathbb{R}^{768}$ and degradation embedding $P_D \in \mathbb{R}^{256}$. The semantic branch $\mathcal{B}_s$ consists of three MLP layers with layer normalization and LeakyReLU activation (except for the last layer). The semantic branch serves two purposes: first, to generate a degradation-independent semantic representation by excluding the degradation information contained in $P_{CLIP}$, and second, to align with the text embedding commonly used in the T2I diffusion model. The degradation branch $\mathcal{B}_d$  consists of two MLP layers with layer normalization and LeakyReLU activation (except for the last layer) to identify the degradation information in $P_{CLIP}$ and assist the image restoration process. 

To ensure that the degradation branch effectively learns the degradation information, we draw inspiration from \cite{conde2024instructir} and design a degradation-aware guidance loss to guide the network in learning useful degradation information. Specifically, we design a simple MLP layer as a classification network $\mathcal{C}$ with the degradation type as the target, such that $d = \mathcal{C}(E_d)$, where $d \in \mathbb{R}^{N}$ and $N$ represents the number of degradation types. We utilize the cross-entropy loss of the classification network $\mathcal{L}_{deg}$ as the degradation-aware guidance loss during training.

\subsection{Image-guided Control Module}
Our primary objective is to utilize the pre-trained T2I diffusion model like Stable Diffusion to guide image restoration. Unlike image generation, image restoration requires maintaining consistent content with the input image. Inspired by the success of ControlNet~\cite{zhang2023controlnet}, we design an additional Image-guided Control Module that provides control signals to the intermediate features of SD, gradually introducing spatial control through zero convolutions, as illustrated in Fig. \ref{fig: overview}. 

Our control module consists of a Degradation-aware Control Encoder, a Control Decoder (CD), and a Image Control Network. Directly using the architecture of the ControlNet can introduce control signals with degradation and undesirable color, possibly due to the inability of the ControlNet architecture to adaptively handle different degradation types. Hence, we design a Degradation-aware Control Encoder that encodes the low-quality image and the degradation embedding $P_D$ to differentiate between different degradation tasks. To ensure meaningful information can be encoded, we also employ a reconstruction loss $\mathcal{L}_{rec}$ to supervise the control reconstruction obtained by our Control Decoder. 

\begin{figure}[t] 
\centering
\includegraphics[width=0.32\textwidth]{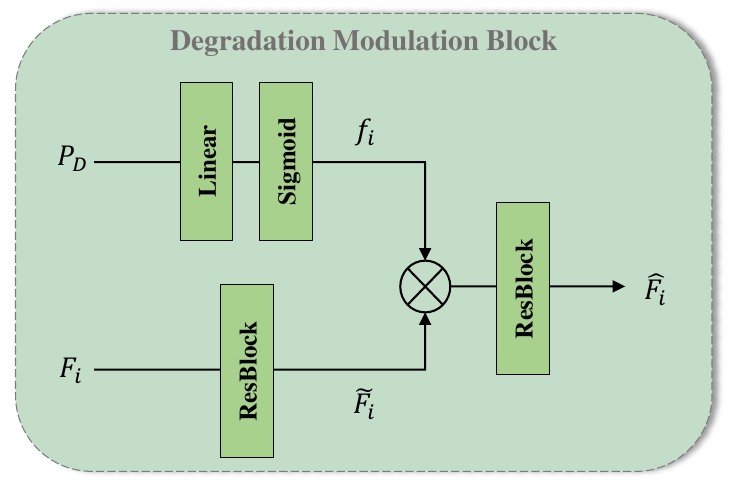}
\caption{The architecture of the proposed Degradation Modulation Block (DMB) in the Control Encoder.}
\vspace{-3mm}
\label{fig: dmb}
\end{figure}

Our Control Encoder consists of four residual blocks with Degradation Modulation Blocks (DMB). The design of our Degradation Modulation Block is illustrated in Fig. \ref{fig: dmb}. For the $i_{th}$ block, feature $F_i$ is first transformed into $\tilde{F_i}$ through the residual block. Given the degradation embedding $P_D$, it passes through a linear layer and a sigmoid activation function to generate a modulation vector $f_i$. The modulation vector is then used for channel-wise multiplication $\otimes$ with image feature $\tilde{F_i}$ to select degradation task-related feature. Last, the output feature is generated from the residual block. The process is summarized as follows:
\begin{equation}
\begin{aligned}
   \tilde{F_i} &= \texttt{ResBlock}(F_i), \\
   f_i &= \texttt{Sigmoid}(\texttt{Linear}(P_D), \\
   \hat{F_i} &= \texttt{ResBlock}(\tilde{F_i} \otimes f_i).
\end{aligned}
\label{eq3}
\end{equation}

The Control Decoder consists of four residual blocks to reconstruct the image $I_{CD}$ for supervision. During the inference process, the Control Decoder is not required since we do not need supervision. The $L_{rec}$ is the L2 loss of the reconstruction $I_{CD}$ and the ground truth $I_{GT}$:
\begin{equation}
    \mathcal{L}_{rec} = ||I_{CD}-I_{GT}||_2.
\label{eq4}
\end{equation}

\subsection{Degradation-aware Decoder}
The pre-trained SD adopts the VAE to compress the images into latent codes for diffusion and reduce computational costs. However, due to the compression of VAE, employing the original SD VAE's decoder to decode the latent codes can result in certain issues of details distortion\cite{zhu2023asymmetric}, such as in facial features and textual content, as depicted in Fig. \ref{fig: ab_decoder}.

\begin{figure}[t] 
\centering
\includegraphics[width=0.48\textwidth]{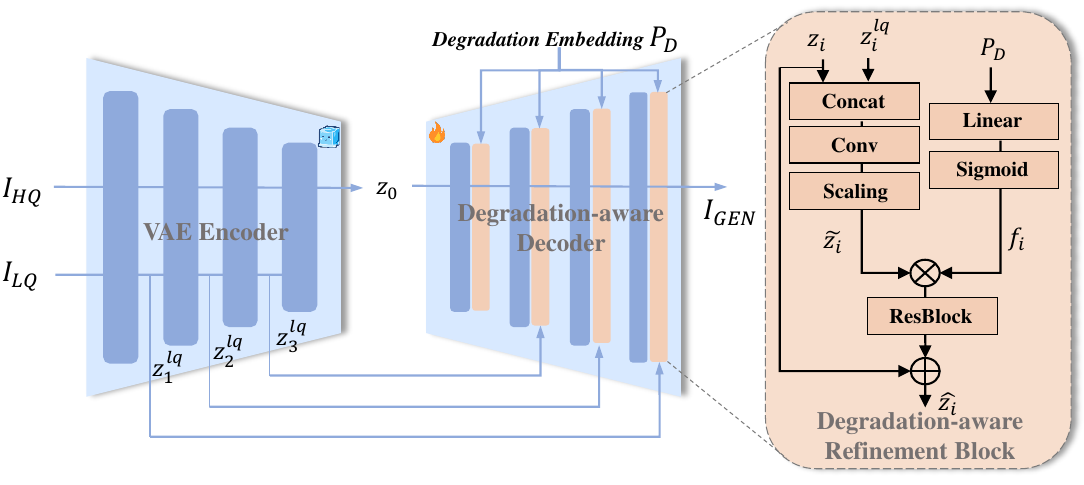}
\vspace{-2mm}
\caption{The training strategy of our Degradation-aware Decoder and the architecture of the proposed Degradation-aware Refinement Block (DRB).}
\vspace{-3mm}
\label{fig: decoder}
\end{figure}

To address this issue, we propose a Degradation-aware Decoder to decode the latent codes for image reconstruction. Specifically, we utilize the intermediate features $z_{1}^{lq}$, $z_{2}^{lq}$, and $z_{3}^{lq}$ extracted from the LQ image by the VAE encoder for assisting in detail refinement. As our method needs to adapt to different degradation types, the decoder should possess degradation awareness. Therefore, we introduce a Degradation-aware Refinement Block with the guidance of degradation embedding $P_D$ and devise an effective training strategy. As shown in Fig. \ref{fig: decoder}, during training, we use the VAE encoder to generate $z_0$ from $I_{HQ}$, while $z_0$ is generated from the diffusion process during the inference time. The decoder decodes $z_0$ into the image domain, combining the LQ features $z_{1}^{lq}$, $z_{2}^{lq}$, and $z_{3}^{lq}$, and the degradation embedding $P_D$. To achieve the combination of this information, we design a Degradation-aware Refinement Block (DRB). Specifically, the decoder features $z_i$ are first concatenated with the encoder features $z_{i}^{lq}$, and then convolved to obtain refined features $\tilde{z_i}$. This process enables direct information transformation from the encoder to the decoder. Then, in order to utilize the degradation embedding $P_D$ for control, we employ similar operations as the Image-guided Control Module. The degradation embedding $P_D$ undergoes a linear layer and a sigmoid activation function to generate a modulation vector $f_i$. $f_i$ is then used to perform channel-wise multiplication $\otimes$ with the combined features $\tilde{z_i}$ for feature selection. Finally, the features $\hat{z_i}$ are corrected through residual blocks. The formulation of this process is as follows:
\begin{equation}
\begin{aligned}
   \tilde{z_i} &= \texttt{Conv}(\texttt{Concat}(z_i, z_{i}^{lq})), \\
   f_i &= \texttt{Sigmoid}(\texttt{Linear}(P_D), \\
   \hat{z_i} &= \texttt{ResBlock}(\tilde{z_i} \otimes f_i) + z_{i}.
\end{aligned}
\label{eq5}
\end{equation}

\subsection{Loss Function}
Our training process consists of two stages: the first stage trains the conditioned diffusion model, and the second stage trains the Degradation-aware Decoder. During the training of the first stage, the HQ image is encoded by the pre-trained VAE encoder\cite{rombach2022ldm} to obtain the latent code $z_0$. The diffusion process progressively adds noise to $z_0$ to generate $z_t$, where $t$ represents the random-sampled diffusion steps. By controlling the diffusion step $t$ and the LQ image $I_{LQ}$, we train the diffusion model, denoted as $\epsilon_{\theta}$, to estimate the noise added to the noise layer $z_t$. The optimization objective of the diffusion process is:

\begin{equation}
\mathcal{L}_{diff} = \mathbbm{E}_{z_0,t,I_{LQ},\epsilon \sim \mathcal{N}}\left[||\epsilon-\epsilon_{\theta}(z_t,t,I_{LQ})||_2^2\right].
\label{eq6}
\end{equation}
The total loss function of the first stage is:
\begin{equation}
    \mathcal{L} = \mathcal{L}_{diff} + \lambda_1\mathcal{L}_{deg} + \lambda_2\mathcal{L}_{rec},
\label{eq7}
\end{equation}
where $\lambda_1$ and $\lambda_2$ are balancing parameters, and we set both $\lambda_{1}$ and $\lambda_{2}$ equal to 1 empirically.

In the decoder training stage, we keep the VAE encoder frozen and only update the parameters of the decoder. Following \cite{zhu2023asymmetric}, the loss function includes reconstruction loss $\mathcal{L}_{gen}$, perceptual loss $\mathcal{L}_{per}$, and adversarial loss $\mathcal{L}_{adv}$. $\mathcal{L}_{gen}$ is the L2 loss of the reconstruction $I_{GEN}$ and the ground truth $I_{GT}$. $\mathcal{L}_{per}$ is calculated on the VGG-19 \cite{simonyan2014vgg} backbone. The total objective for the decoder training is
\begin{equation}
    \mathcal{L}_{dec} = \mathcal{L}_{gen} + \lambda_3\mathcal{L}_{per} + \lambda_4\mathcal{L}_{adv},
\label{eq8}
\end{equation}
where $\lambda_3$ and $\lambda_4$ are balancing parameters and we set $\lambda_{3}$ to be 0.1 and $\lambda_4$ to be 0.001 empirically.

\begin{figure*}[t] 
\centering
\includegraphics[width=\textwidth]{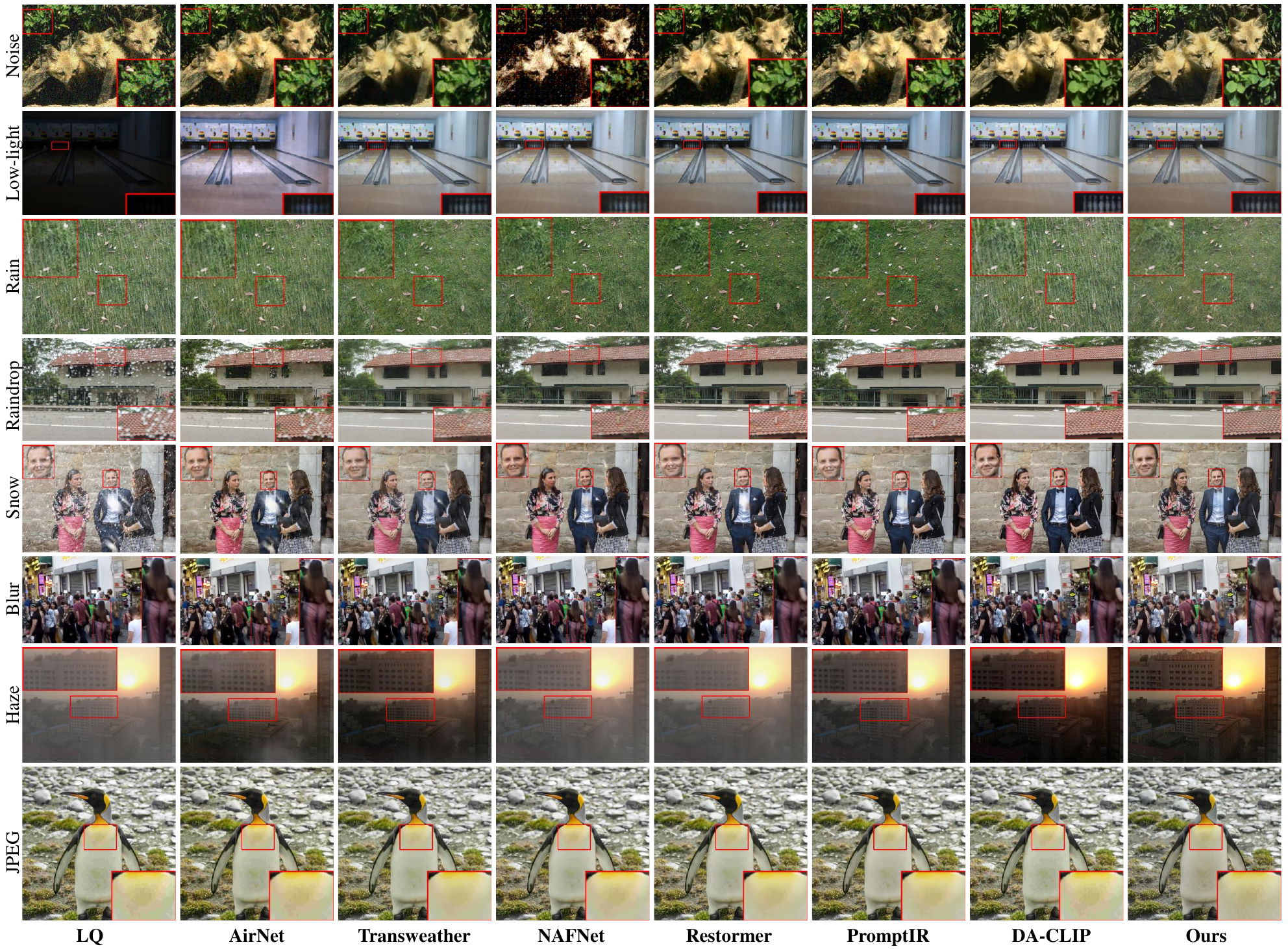}
\vspace{-5mm}
\caption{Qualitative comparison of different methods on the eight single degradation restoration tasks. Our Diff-Restorer notably surpasses others in performance. Magnified regions are provided for clarity and \textbf{please zoom in for the best view.}}
\vspace{-5mm}
\label{fig: compare_8}
\end{figure*}

\begin{figure*}[t] 
\centering
\includegraphics[width=\textwidth]{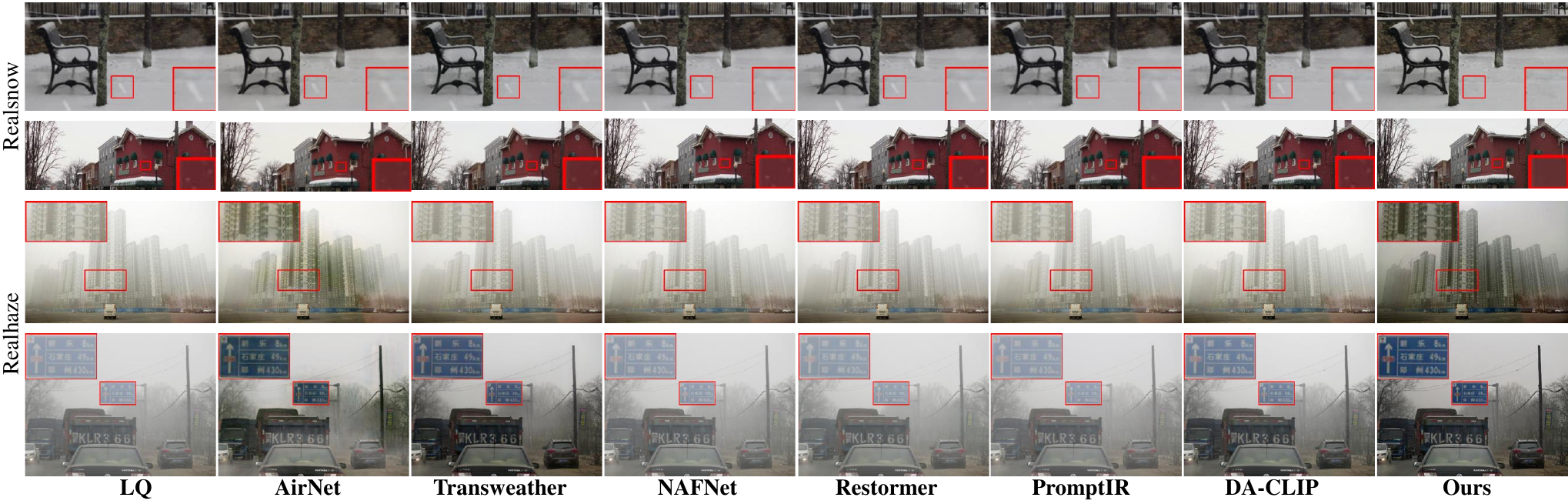}
\vspace{-5mm}
\caption{Qualitative comparison of different methods on real-world restoration tasks. Magnified regions are provided for clarity and \textbf{please zoom in for the best view.}}
\vspace{-5mm}
\label{fig: compare_r}
\end{figure*}

\begin{figure*}[t] 
\centering
\includegraphics[width=\textwidth]{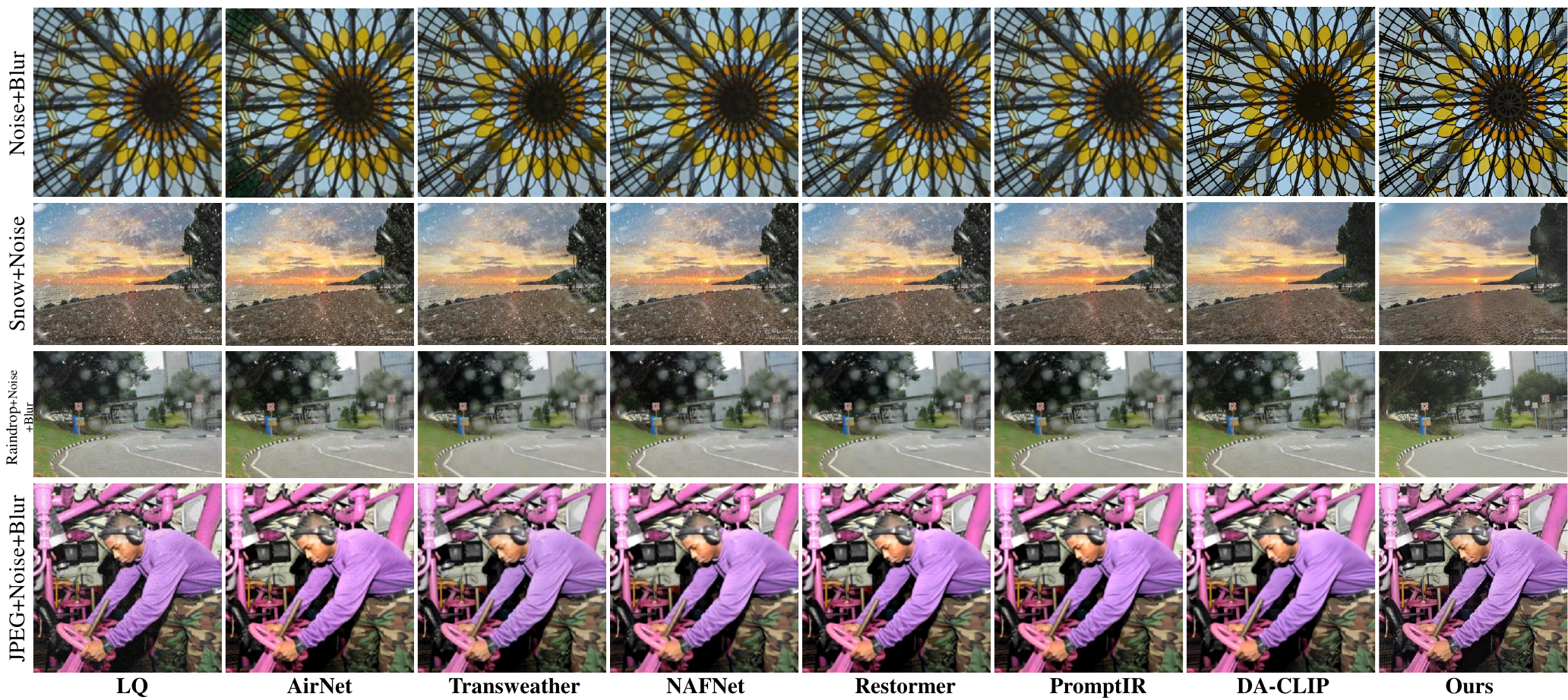}
\vspace{-5mm}
\caption{Qualitative comparison of different methods on mixed degradation restoration tasks. \textbf{Please zoom in for the best view.}}
\vspace{-5mm}
\label{fig: compare_m}
\end{figure*}

\begin{table*}[]
    \caption{Quantitative comparison. \rf{Red} and \bd{blue} colors represent the best and second best performance. $\downarrow$ represents the smaller the better, and $\uparrow$ represents the bigger the better. Ours$_{S}$ represents our method with SD decoder.}
    \vspace{-2mm}
    \centering
    \resizebox{1.0\textwidth}{!}{
    \begin{tabular}{p{1.3cm}p{1.7cm}cccccc||p{1.3cm}p{1.7cm}cccccc}
    \toprule
        Degradation & Method & FID$\downarrow$ & LPIPS$\downarrow$ & DISTS$\downarrow$ & NIQE$\downarrow$ & MUSIQ$\uparrow$ & CLIPIQA$\uparrow$ & Degradation & Method & FID$\downarrow$ & LPIPS$\downarrow$ & DISTS$\downarrow$ & NIQE$\downarrow$ & MUSIQ$\uparrow$ & CLIPIQA$\uparrow$ \\
        \midrule
        \multirow{8}{*}{\begin{minipage}{1.3cm}Noise\end{minipage}} 
        & AirNet~\cite{li2022airnet} & 56.07 & 0.2104 & 0.1805 & \rf{2.82} & 60.99 & 0.4923 & \multirow{8}{*}{\begin{minipage}{1.3cm}Low-light \end{minipage}} 
        & AirNet~\cite{li2022airnet} & 166.31 & 0.3047 & 0.2155 & \bd{3.90} & 54.31 & 0.2891 \\
        & TransWeather~\cite{valanarasu2022transweather} & \bd{43.50} & 0.1519 & 0.1524 & 4.77 & 63.42 & 0.5486 &
        & TransWeather~\cite{valanarasu2022transweather} & 58.98 & 0.1283 & 0.1094 & 4.17 & 69.35 & 0.4294 \\
        & NAFNet~\cite{chen2022nafnet} & 100.90 & 0.3216 & 0.2104 & 3.76 & 58.01 & 0.4863 &
        & NAFNet~\cite{chen2022nafnet} & 88.53 & 0.2010 & 0.1662 & \rf{3.79} & 56.04 & 0.3089 \\
        & Restormer~\cite{zamir2022restormer} & 46.62 & \bd{0.1484} & 0.1480 & 3.55 & 66.24 & 0.5928 &
        & Restormer~\cite{zamir2022restormer} & 73.61 & 0.1840 & 0.1438 & 4.36 & 65.49 & 0.3553 \\
        & PromptIR~\cite{potlapalli2024promptir} & 48.29 & 0.1797 & 0.1534 & \bd{3.32} & 67.20 & 0.5848 & 
        & PromptIR~\cite{potlapalli2024promptir} & 63.62 & 0.1371 & 0.1154 & 4.38 & 69.63 & 0.4220 \\
        & DA-CLIP~\cite{luo2023da-clip} & 77.91 & 0.2023 & 0.1829 & 4.99 & 68.45 & 0.5342 &
        & DA-CLIP~\cite{luo2023da-clip} & \rf{34.63}  & \rf{0.0800} & \rf{0.0769} & 4.70  & \bd{73.76}  & \bd{0.4839} \\
        & Ours$_{S}$ & 47.62  & 0.2201 & \bd{0.1401} & 4.04  & \rf{71.48}  & \rf{0.7616}  & 
        & Ours$_{S}$ & 52.60 & 0.1415 & 0.1016 & 4.01 & \rf{73.80} & \rf{0.5796} \\
        & Ours & \rf{35.32}  & \rf{0.1413} & \rf{0.1075} & 3.50  & \bd{69.53}  & \bd{0.6540} &
        & Ours & \bd{40.23}  & \bd{0.1105} & \bd{0.0873} & 4.42  & 71.69  & 0.3970 \\
        \midrule
        \multirow{8}{*}{\begin{minipage}{1.3cm}Haze\end{minipage}} 
        & AirNet~\cite{li2022airnet} &  26.40  & 0.0714 & 0.0741 & \bd{3.28}  & 62.61  & 0.4267  & \multirow{8}{*}{\begin{minipage}{1.3cm}Rain \end{minipage}} 
        & AirNet~\cite{li2022airnet} & 71.79  & 0.1641 & 0.1530 & 3.77  & 65.15  & 0.5428 \\
        & TransWeather~\cite{valanarasu2022transweather} & \bd{7.68}   & \bd{0.0284} & 0.0369 & 3.43  & 62.89  & 0.3866   &
        & TransWeather~\cite{valanarasu2022transweather}  & 26.61  & 0.0639 & 0.0776 & 3.67  & 68.67  & 0.6326 \\
        & NAFNet~\cite{chen2022nafnet} & 13.41  & 0.0489 & 0.0507 & 3.36  & 61.12  & 0.3478 &
        & NAFNet~\cite{chen2022nafnet} & 37.87  & 0.0956 & 0.0994 & 3.35  & 67.75  & 0.6042 \\
        & Restormer~\cite{zamir2022restormer} & 14.71  & 0.0450 & 0.0565 & 3.48  & 61.41  & 0.3633 &
        & Restormer~\cite{zamir2022restormer} & \rf{23.59}  & \bd{0.0619} & \rf{0.0638} & 3.34  & 68.58  & 0.0638 \\
        & PromptIR~\cite{potlapalli2024promptir} & 8.64  & 0.0290 & \bd{0.0359} & 3.40  & 61.58  & 0.3690 &
        & PromptIR~\cite{potlapalli2024promptir} & 25.25  & \rf{0.0580} & 0.0701 & \bd{3.20}  & \bd{69.54}  & \bd{0.6779} \\
        & DA-CLIP~\cite{luo2023da-clip} & \rf{4.17} & \rf{0.0157} & \rf{0.0249} & 3.55  & 61.34  & 0.3763 &
        & DA-CLIP~\cite{luo2023da-clip} & 95.15  & 0.1339 & 0.1113 & 3.43  & 67.47  & 0.6508 \\
        & Ours$_{S}$ & 27.11  & 0.1657 & 0.1102 & 3.52  & \rf{69.52}  & \rf{0.5440} &
        & Ours$_{S}$  & 44.11  & 0.1605 & 0.1029 & 3.29  & \rf{69.63}  & \rf{0.6821} \\
        & Ours & 12.13  & 0.0670 & 0.0625 & \rf{3.23}  & \bd{63.92}  & \bd{0.4777} &
        & Ours & \bd{24.45}  & 0.0840 & \bd{0.0683} & \rf{3.02}  & 67.37  & 0.6390 \\
        \midrule
        \multirow{8}{*}{\begin{minipage}{1.3cm}Raindrop\end{minipage}} 
        & AirNet~\cite{li2022airnet} & 57.25  & 0.1736 & 0.1227 & \bd{3.11}  & 68.21  & 0.3883   &  \multirow{8}{*}{\begin{minipage}{1.3cm}Snow \end{minipage}} 
        & AirNet~\cite{li2022airnet} & 63.00  & 0.1630 & 0.1280 & 3.00  & 65.06  & 0.3480 \\
        & TransWeather~\cite{valanarasu2022transweather}  & 31.04  & 0.1007 & 0.7957 & 3.33  & 69.47  & 0.4567 &
        & TransWeather~\cite{valanarasu2022transweather} & 35.24 & 0.0954 & 0.0824 & 3.04 & 69.50 & 0.4240 \\
        & NAFNet~\cite{chen2022nafnet} & 32.58  & 0.1148 & 0.0888 & 3.14  & 68.80  & 0.4046   &
        & NAFNet~\cite{chen2022nafnet} & 35.36  & 0.0985 & 0.0860 & 2.99  & 67.89  & 0.4049 \\
        & Restormer~\cite{zamir2022restormer} & 28.82  & 0.1144 & 0.0896 & 3.29  & 70.71  & 0.4207   &
        & Restormer~\cite{zamir2022restormer} & 29.36  & \bd{0.0808} & 0.0678 & 3.21  & \bd{70.74}  & 0.4567 \\
        & PromptIR~\cite{potlapalli2024promptir}  & 24.45  & \bd{0.0988} & \bd{0.0776} & 3.27  & 70.83  & 0.4331   &
        & PromptIR~\cite{potlapalli2024promptir} & 32.34  & \rf{0.0775} & 0.0783 & 3.31  & \rf{71.58}  & \rf{0.5346}\\
        & DA-CLIP~\cite{luo2023da-clip} & \rf{16.55}  & \rf{0.0771} & \rf{0.0548} & 3.55  & 70.65  & 0.5261   &
        & DA-CLIP~\cite{luo2023da-clip} & \bd{24.36}  & 0.0897 & \bd{0.0675} & \bd{2.96}  & 70.52  & 0.4555\\
        & Ours$_{S}$ & 32.04  & 0.1782 & 0.1130 & 3.32  & \rf{72.74}  & \rf{0.6245}   &
        & Ours$_{S}$ & 33.03  & 0.1509 & 0.0934 & 3.11  & 70.58  & \bd{0.4798} \\
        & Ours & \bd{22.50}  & 0.1282 & 0.0936 & \rf{3.10}  & \bd{70.86}  & \bd{0.5299}   &
        & Ours & \rf{23.84}  & 0.0840 & \rf{0.0650} & \rf{2.93}  & 69.41  & 0.4436 \\
        \midrule
        \multirow{8}{*}{\begin{minipage}{1.3cm}Blur\end{minipage}} 
        & AirNet~\cite{li2022airnet} & 44.65  & 0.2931 & 0.1631 & 5.85  & 27.90  & 0.2111   & \multirow{8}{*}{\begin{minipage}{1.3cm}JPEG Compression Artifact \end{minipage}} 
        & AirNet~\cite{li2022airnet} & 37.75  & 0.1339 & 0.1230 & \bd{3.64}  & 67.82  & 0.5103 \\
        & TransWeather~\cite{valanarasu2022transweather}  & 40.92  & 0.2404 & 0.1594 & 5.48  & 31.19  & 0.2093   &
        & TransWeather~\cite{valanarasu2022transweather} & \bd{29.22}  & \bd{0.1012} & \bd{0.0974} & \rf{3.60}  & 69.09  & 0.4999 \\
        & NAFNet~\cite{chen2022nafnet} & 35.60  & 0.2653 & 0.1466 & 5.77  & 29.90  & 0.2004   & 
        & NAFNet~\cite{chen2022nafnet} & 32.66  & 0.1122 & 0.1107 & 3.95  & 69.58  & 0.5211 \\
        & Restormer~\cite{zamir2022restormer} & 39.45  & 0.2849 & 0.1649 & 6.04  & 29.41  & 0.2064   &
        & Restormer~\cite{zamir2022restormer} & 30.52  & 0.1028 & 0.1100 & 4.20  & 70.41  & 0.5616 \\
        & PromptIR~\cite{potlapalli2024promptir} & \bd{21.91}  & 0.1877 & 0.1120 & 5.46  & 36.98  & 0.2266   & 
        & PromptIR~\cite{potlapalli2024promptir} & 30.49  & 0.1040 & 0.1113 & 4.23  & 70.74  & 0.5689 \\
        & DA-CLIP~\cite{luo2023da-clip} & \rf{13.82}  & \rf{0.1271} & \rf{0.0752} & 4.50  & 40.52  & 0.2153   &
        & DA-CLIP~\cite{luo2023da-clip} & 32.97  & 0.1217 & 0.1213 & 4.69  & \rf{73.21}  & 0.6602 \\
        & Ours$_{S}$  & 30.72  & 0.2391 & 0.1548 & \rf{3.14}  & \rf{54.49}  & \rf{0.3596}   &
        & Ours$_{S}$ & 37.61  & 0.1402 & 0.1013 & 3.96  & \bd{72.41}  & \rf{0.7535} \\
        & Ours  & 26.05  & \bd{0.1896} & \bd{0.1074} & \bd{3.93}  & \bd{46.90}  & \bd{0.2332}   &
        & Ours & \rf{24.55}  & \rf{0.0840} & \rf{0.0923} & 3.86  & 71.27  & \bd{0.6825} \\
        \midrule
        \multirow{8}{*}{\begin{minipage}{1.3cm}Real-world:\\ Snow \end{minipage}} 
        & AirNet~\cite{li2022airnet} & 93.08  & 0.1626 & 0.1277 & 4.17  & 52.87  & 0.4301   & \multirow{8}{*}{\begin{minipage}{1.3cm}Real-world:\\ Haze \end{minipage}} 
        & AirNet~\cite{li2022airnet} & - & -      & -      & 4.65  & 56.17  & 0.4103 \\
        & TransWeather~\cite{valanarasu2022transweather} & 76.03  & 0.1583 & 0.1235 & 4.25  & \bd{55.70}  & \bd{0.4891}   &
        & TransWeather~\cite{valanarasu2022transweather} & - & -      & -      & 5.05  & 57.48  & 0.3819 \\
        & NAFNet~\cite{chen2022nafnet} & 67.77  & \bd{0.1225} & 0.1087 & 4.16  & 53.95  & 0.4373   & 
        & NAFNet~\cite{chen2022nafnet} & - & -      & -      & 5.29  & 56.18  & 0.3854 \\
        & Restormer~\cite{zamir2022restormer} & 72.59  & 0.1296 & 0.1123 & 4.38  & 55.36  & 0.4585   & 
        & Restormer~\cite{zamir2022restormer} & - & -      & -      & 5.80  & 57.22  & 0.4006 \\
        & PromptIR~\cite{potlapalli2024promptir}  & 67.53  & 0.1238 & \rf{0.1011} & 4.28  & 54.58  & 0.4711   & 
        & PromptIR~\cite{potlapalli2024promptir}  & - & -      & -      & 5.29  & 56.61  & 0.4089 \\
        & DA-CLIP~\cite{luo2023da-clip} & \bd{61.61}  & \rf{0.1222} & 0.1059 & \rf{4.08}  & 52.76  & 0.4662   &
        & DA-CLIP~\cite{luo2023da-clip} & - & -      & -      & 4.71  & 55.47  & 0.4052 \\
        & Ours$_{S}$ & 100.41 & 0.2261 & 0.1633 & 4.51  & \rf{55.81}  & \rf{0.4990}   &
        & Ours$_{S}$ & - & -      & -      & \bd{4.34}  & \rf{62.93}  & \rf{0.4834} \\
        & Ours & \rf{60.92}  & 0.1260 & \bd{0.1042} & \bd{4.11}  & 54.54  & 0.4685   &
        & Ours & - & -      & -      & \rf{4.27}  & \bd{59.93}  & \bd{0.4369} \\
        \midrule
        \multirow{8}{*}{\begin{minipage}{1.3cm}Mixture\\(N=2):\\Blur$+$\\Noise \end{minipage}} & AirNet~\cite{li2022airnet} & 75.70  & 0.4508 & 0.2500 & 6.87  & 41.24  & 0.3212  & \multirow{8}{*}{\begin{minipage}{1.3cm}Mixture\\(N=2): \\Snow$+$\\Noise \end{minipage}}
        & AirNet~\cite{li2022airnet} & 113.26 & 0.3670 & 0.2291 & 5.17  & 54.91  & 0.2542 \\
        & TransWeather~\cite{valanarasu2022transweather} & 68.32  & 0.4123 & 0.2411 & 7.00  & 39.62  & 0.2567   &
        & TransWeather~\cite{valanarasu2022transweather} & 110.87 & 0.3448 & 0.2154 & 4.76  & 54.92  & 0.3051  \\
        & NAFNet~\cite{chen2022nafnet} & 70.97  & 0.4213 & 0.2406 & 7.07  & 44.68  & 0.2999   &
        & NAFNet~\cite{chen2022nafnet} & 109.20 & 0.3793 & 0.2278 & 5.82  & 53.35  & 0.2827\\
        & Restormer~\cite{zamir2022restormer} & 65.96  & 0.4279 & 0.2363 & 7.03  & 44.33  & 0.3331   &
        & Restormer~\cite{zamir2022restormer} & 110.63 & 0.3670 & 0.2199 & 6.03  & 54.81  & 0.3383 \\
        & PromptIR~\cite{potlapalli2024promptir} & 68.05  & 0.4489 & 0.2412 & 7.30  & 41.41  & 0.3212   &
        & PromptIR~\cite{potlapalli2024promptir} & 77.15  & 0.3281 & 0.1944 & 6.43  & 57.78  & 0.3126\\
        & DA-CLIP~\cite{luo2023da-clip} & 80.66  & 0.4675 & 0.2593 & 7.10  & 37.72  & 0.2995   & 
        & DA-CLIP~\cite{luo2023da-clip}  & 83.51  & 0.3774 & 0.2198 & 5.80  & 53.41  & 0.2722\\
        & Ours$_{S}$ & \bd{37.90}  & \bd{0.1807} & \bd{0.1306} & \rf{4.33}  & \rf{73.87}  & \rf{0.7876}   & 
        & Ours$_{S}$ & \bd{39.18}  & \bd{0.1897} & \bd{0.1113} & \rf{2.86}  & \rf{69.47}  & \rf{0.4616} \\
        & Ours & \rf{26.59}  & \rf{0.1438} & \rf{0.1240} & \bd{4.48}  & \bd{72.55}  & \bd{0.5864}   &
        & Ours & \rf{33.41}  & \rf{0.1435} & \rf{0.0949} & \bd{2.88}  & \bd{68.19}  & \bd{0.4022} \\
        \midrule
        \multirow{8}{*}{\begin{minipage}{1.3cm}Mixture\\(N=3):\\Blur$+$\\Noise$+$\\Raindrop\end{minipage}} 
         & AirNet~\cite{li2022airnet} & 164.10 & 0.5096 & 0.2983 & 4.77  & 24.27  & 0.1535   & \multirow{8}{*}{\begin{minipage}{1.3cm}Mixture\\(N=3):\\Blur$+$\\Noise$+$\\JPEG\end{minipage}}
        & AirNet~\cite{li2022airnet} &  81.44  & 0.3986 & 0.2647 & 5.05  & 36.06  & 0.2702 \\
        & TransWeather~\cite{valanarasu2022transweather} & 167.38 & 0.5059 & 0.2811 & 7.73  & 26.61  & 0.1522 &
        & TransWeather~\cite{valanarasu2022transweather} & 73.45  & 0.3911 & 0.2514 & 7.50  & 37.96  & 0.2825  \\
        & NAFNet~\cite{chen2022nafnet} & 171.12 & 0.5636 & 0.2967 & 5.38  & 25.21  & 0.1707   &
        & NAFNet~\cite{chen2022nafnet} & 82.06  & 0.4406 & 0.2625 & 5.63  & 36.28  & 0.2716\\
        & Restormer~\cite{zamir2022restormer} & 189.01 & 0.5350 & 0.2883 & 6.91  & 30.84  & 0.2079   & 
        & Restormer~\cite{zamir2022restormer} & 88.61  & 0.4104 & 0.2472 & 6.50  & 41.78  & 0.3231 \\
        & PromptIR~\cite{potlapalli2024promptir} & 179.66 & 0.5169 & 0.2762 & 6.56  & 31.68  & 0.2096   &
        & PromptIR~\cite{potlapalli2024promptir} & 85.71  & 0.4088 & 0.2451 & 6.51  & 41.72  & 0.3384\\
        & DA-CLIP~\cite{luo2023da-clip} & 178.20 & 0.5571 & 0.3194 & 9.19  & 26.11  & 0.1907   &
        & DA-CLIP~\cite{luo2023da-clip} & 102.72 & 0.4464 & 0.2763 & 7.74  & 37.27  & 0.3110 \\
        & Ours$_{S}$ & \bd{79.03}  & \bd{0.2627} & \rf{0.1507} & \rf{4.04}  & \rf{68.62}  & \rf{0.5816}   &
        & Ours$_{S}$ & \bd{59.93}  & \bd{0.2352} & \rf{0.1484} & \bd{4.23}  & \rf{69.78}  & \rf{0.7109} \\
        & Ours & \rf{74.24}  & \rf{0.2511} & \bd{0.1571} & \bd{4.07}  & \bd{61.91}  & \bd{0.3851}   &
        & Ours & \rf{52.30}  & \rf{0.2141} & \bd{0.1555} & \rf{4.04}  & \bd{65.00}  & \bd{0.4847} \\
        \bottomrule
    \end{tabular}}
    \vspace{-3mm}
    \label{tab: compare}
\end{table*}

\section{Experiments}
\subsection{Experimental Settings}
\textbf{Datasets.} To train the Diff-Restorer, we considered a total of 8 degradation types: ``Noise", ``Low-light", ``Haze", ``Rain", ``Raindrop", ``Snow", ``Blur", and ``JPEG compression artifact". For each degradation type, we collect data from the corresponding datasets, ``Noise": DIV2K\cite{agustsson2017div2k} and Flickr2K\cite{timofte2017flickr2k}, ``Low-light": LOL\cite{wei2018lol}, ``Haze": Reside\cite{li2018reside}, ``Rain": Rain1400\cite{fu2017rain1400}, ``Raindrop": Raindrop\cite{qian2018raindrop}, ``Snow": Snow100K\cite{liu2018snow100k}, ``Blur": Gopro\cite{nah2017gopro}, and ``JPEG compression artifact": DIV2K\cite{agustsson2017div2k} and Flickr2K\cite{timofte2017flickr2k}. The noisy image is generated by adding Gaussian noise with noise level 50 and the JPEG-compressing images are synthetic data with a JPEG quality factor of 10. What's more, due to the unequal number of images in each dataset, for larger datasets such as Snow100K\cite{liu2018snow100k}, we randomly sampled 10K images from them and for smaller datasets like LOL\cite{wei2018lol}, we oversampled them by performing data augmentation techniques such as rotation, affine transformation, noise addition, and random cropping. As a result, we unify the sample size to 10K for all datasets, and all images are resized to 512x512 for training. Thus, we obtain a large dataset with multiple degradation types. For testing, we conducted evaluations on both single degradation datasets and mixed degradation datasets. For single degradation datasets, we used the corresponding test sets associated with the training sets for evaluation. What's more, to evaluate the method's generality, we also make evaluations on real-world datasets: RealSnow\cite{zhu2023realsnow} and RTTS\cite{li2018reside}. For mixed degradation datasets, we primarily focused on the mixed degradations with two and three degradations and selected representative restoration tasks to construct the mixed degradation datasets for testing. The degradations of our test sets include ``Noise$+$Blur", ``Snow$+$Noise", ``Raindrop$+$Noise$+$Blur", and ``JPEG$+$Noise$+$Blur".

\textbf{Evaluation Metrics.}
To provide a comprehensive and effective quantitative evaluation of the different methods, we employ a range of widely used metrics, including FID\cite{heusel2017fid},  LPIPS\cite{zhang2018lpips}, DISTS\cite{ding2020dists}, NIQE\cite{zhang2015niqe}, MUSIQ\cite{ke2021musiq}, and CLIPIQA\cite{wang2023clipiqa}. It is worth noting that we did not utilize pixel-level metrics such as PSNR and SSIM\cite{wang2004ssim} because these metrics can not fully reflect the perceptual image quality as perceived by humans as illustrated in \cite{yu2024supir}. FID\cite{heusel2017fid}, LPIPS\cite{zhang2018lpips} and DISTS\cite{ding2020dists} are reference-based perceptual metrics that measure the distance and similarity between ground truth images and the restored images. NIQE\cite{zhang2015niqe}, MUSIQ\cite{ke2021musiq}, and CLIPIQA\cite{wang2023clipiqa} are non-reference perceptual metrics that evaluate the quality of generated images. We employ the IQA-Pytorch \footnote{https://github.com/chaofengc/IQA-PyTorch} to
compute these metrics.

\textbf{Compared Methods.}
We compare our Diff-Restorer with state-of-the-art all-in-one image restoration methods: AirNet\cite{li2022airnet}, TransWeather\cite{valanarasu2022transweather}, NAFNet\cite{chen2022nafnet}, Restormer\cite{zamir2022restormer} and PromptIR\cite{potlapalli2024promptir}. We also compare our method with the diffusion-based method: DA-CLIP\cite{luo2023da-clip}. To make a fair evaluation, we retrain all methods from scratch with our datasets. Since we aim to solve the universal image restoration problem, we do not compare with methods that only consider a single task.

\textbf{Implementation Details.}
We employ the pre-trained SD 1.5 \footnote{https://huggingface.co/runwayml/stable-diffusion-v1-5} model as the base pre-trained model. During training, we finetune the model with AdamW optimizer\cite{loshchilov2017adamw} for 100 epochs. The batch size and the learning rate are set to 16 and 1e-5. All experiments are conducted on two NVIDIA A100 GPUs. The training time is about 150 hours. We use UniPCMultistepScheduler\cite{zhao2024unipc} sampling with 20 timesteps. To train the Degradation-aware Decoder, we finetune the model with AdamW optimizer \cite{loshchilov2017adamw} for 25 epochs. The batch size is 4 and the learning rate is the same as the training of diffusion model. The training time of the decoder is about 75 hours.

\subsection{Experimental Results}
\textbf{Qualitative Results.}
Fig. \ref{fig: compare_8} demonstrates the superior performance of our Diff-Restorer on eight challenging single degradation restoration tasks. It can be observed that our method exhibits significant advantages in detail generation and degradation removal, particularly in the texture of leaves (row 1), facial features (row 5), and the appearance of penguins (row 8). Fig. \ref{fig: compare_r} showcases the effectiveness of our Diff-Restorer in snow and haze removal on real-world datasets compared to other methods. It can be observed that our method outperforms other approaches in the performance of degradation removal. Additionally, although DA-CLIP performs well on the eight restoration tasks shown in Fig. \ref{fig: compare_8}, it falls short in real-world image restoration. Our method, on the other hand, shows great performance on real-world restoration tasks, demonstrating our method's generality. Fig. \ref{fig: compare_m} presents the subjective results of our Diff-Restorer and other methods in handling restoration tasks involving multiple degradations mixed. It can be observed that other methods struggle to handle tasks with mixed degradation removal, especially in the case of ``Raindrop+Noise+Blur" degradation (row 3), where other methods fail to remove raindrops effectively. In contrast, our method achieves excellent restoration results, demonstrating its stability.

\begin{figure*}[ht] 
\centering
\includegraphics[width=0.98\textwidth]{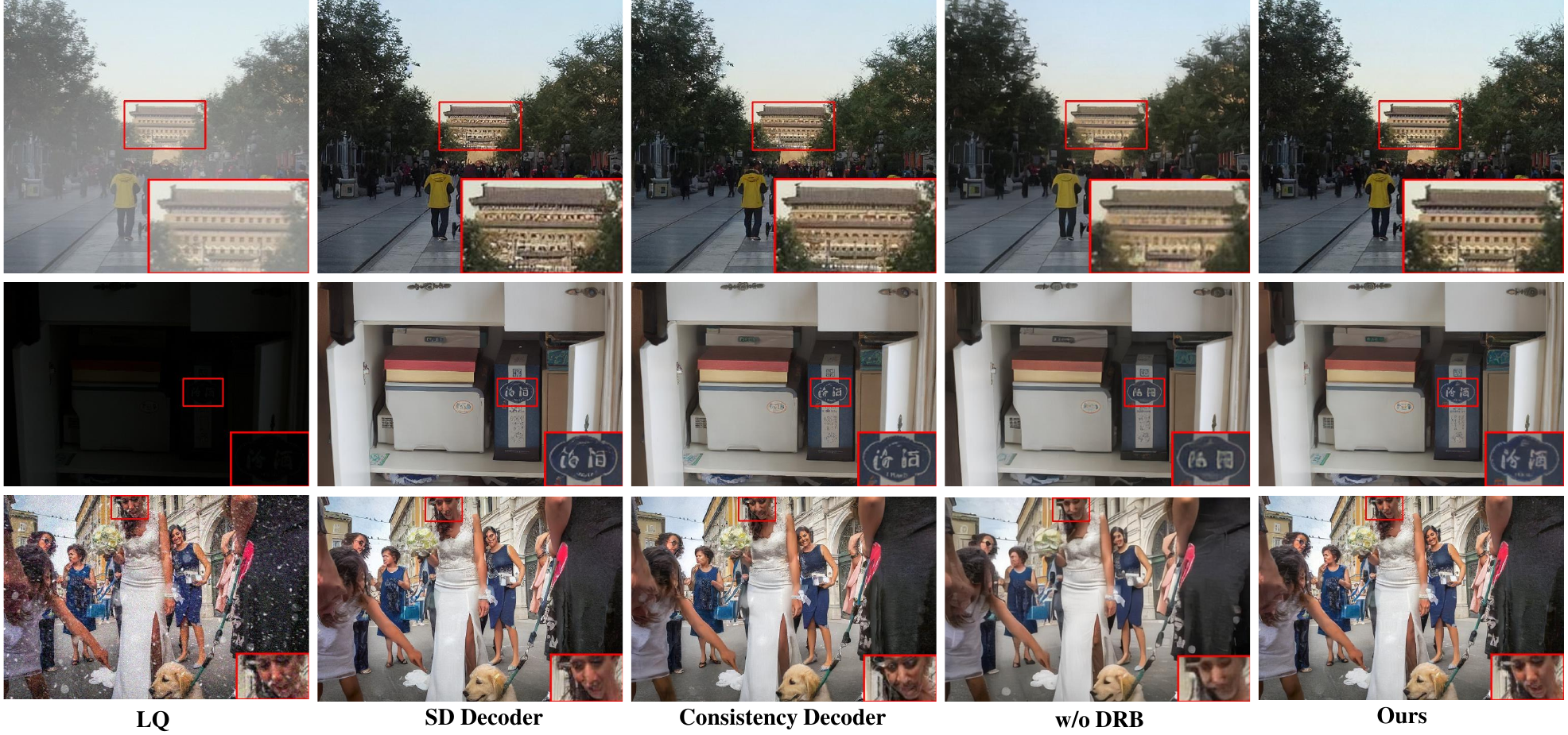}
\vspace{-2mm}
\caption{Qualitative comparison for Diff-Restorer with different VAE decoder. Our decoder exhibits significant advantages in geometric preservation, particularly in the preservation of architectural structures (row 1), text (row 2), and faces (row 3). \textbf{Please zoom in for the best view.}}
\vspace{-3mm}
\label{fig: ab_decoder}
\end{figure*}

\textbf{Quantitative Results.} Table \ref{tab: compare} provides a quantitative comparison between our method and the other six methods on single degradation restoration tasks, real-world restoration tasks, and mixed degradation restoration tasks. We also provide the quantitative results of our method with SD decoder (Ours$_{S}$) to show the effectiveness of the diffusion model in enhancing image perceptual quality. Additionally, since RTTS dataset does not have ground truth, we only calculate the no-reference evaluation metrics for comparison. It can be observed that each method has its advantages in certain restoration tasks. However, our method is the only one that demonstrates good restoration performance across all the tasks. Specifically, our method performs well in denoising and JPEG artifact removal, achieving SOTA performance overall. Additionally, although DA-CLIP performs well on the eight single degradation restoration tasks, it shows average performance in real-world restoration and mixed degradation restoration tasks, which is consistent with the qualitative comparison. Our method exhibits significant advantages in handling real-world restoration and mixed degradation restoration. What's more, while using the SD original decoder achieves better evaluation metrics (particularly MUSIQ and CLIPIQA) than our Degradation-aware Decoder on some restoration tasks, it introduces geometric distortions, especially in text and faces, as shown in the Fig. \ref{fig: ab_decoder}. Therefore, we introduce our decoder for refinement, striking a balance between realism and fidelity. Overall, our method demonstrates competitive performance on most restoration tasks. It is important to note that the primary goal of this paper is not to achieve top-tier performance on all tasks but to drive the development of universal and realistic image restoration models.

\begin{figure}[t] 
\centering
\vspace{-4mm}
\includegraphics[width=0.48\textwidth]{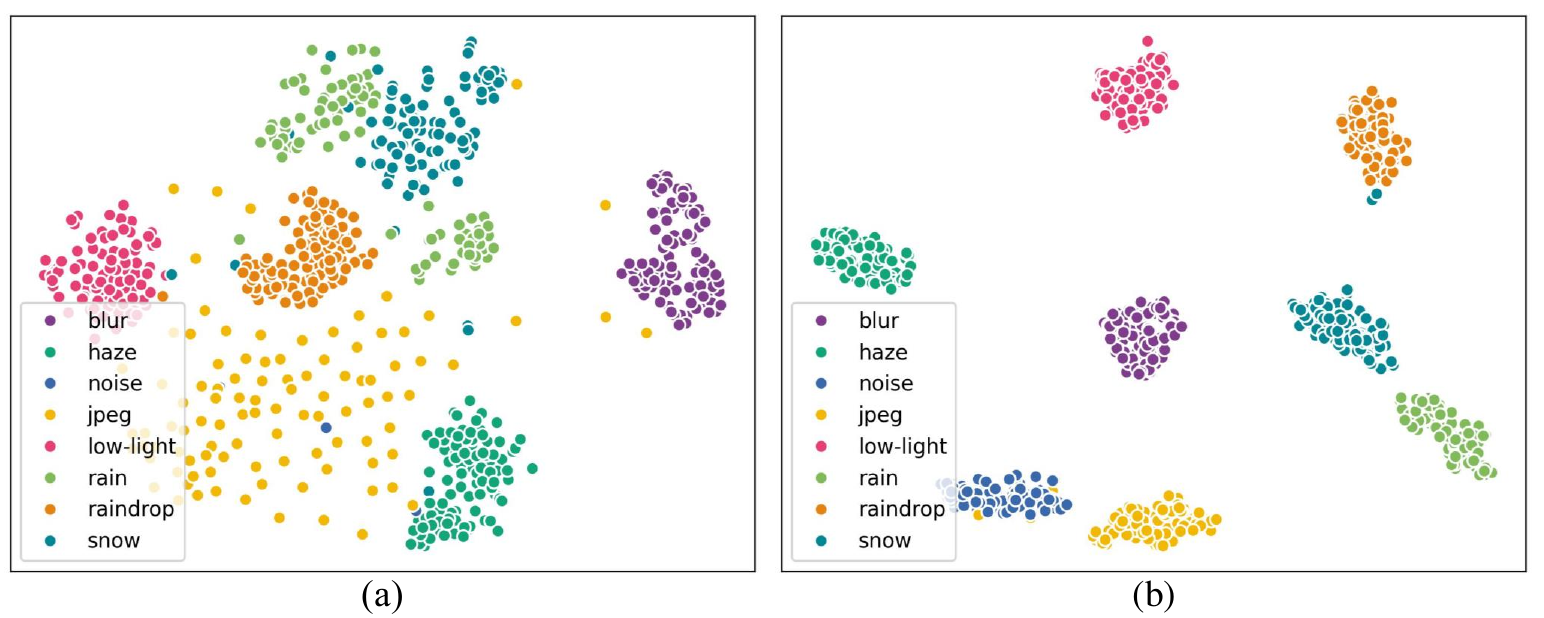}
\caption{t-SNE visualizations of CLIP image embedding and degradation embedding. (a) is the visualization of CLIP image embedding and (b) is the visualization of our degradation embedding.}
\vspace{-7mm}
\label{fig: t-SNE}
\end{figure}

\begin{table}[ht]
\centering
\vspace{-2mm}
\caption{Ablation studies of Visual Prompts. The metrics are reported on the average of the restoration tasks with ``low-light", ``haze", ``realsnow", ``snow+noise" and ``raindrop+blur+noise" degradations.}
\vspace{-2mm}
\label{tab:ab_promptprocessor}
\resizebox{0.40\textwidth}{!}{
\begin{tabular}{lcccc}
\toprule
Method & FID$\downarrow$ & LPIPS$\downarrow$ & DISTS$\downarrow$ & NIQE$\downarrow$ \\ 
\midrule
w/o $\mathcal{B}_s$ & 105.18 & 0.3508 & 0.2436 & 4.55 \\
w/o $P_D$ & 60.30 & 0.2990 & 0.2029  & 4.09\\ 
Ours$_{S}$ & 49.28 & 0.1989 & 0.1308 & 3.71 \\
Ours & 34.23 & 0.1331 & 0.0999 & 3.57 \\ 
\bottomrule
\end{tabular}}
\vspace{-3mm}
\end{table}
\begin{figure}[t] 
\centering
\includegraphics[width=0.50\textwidth]{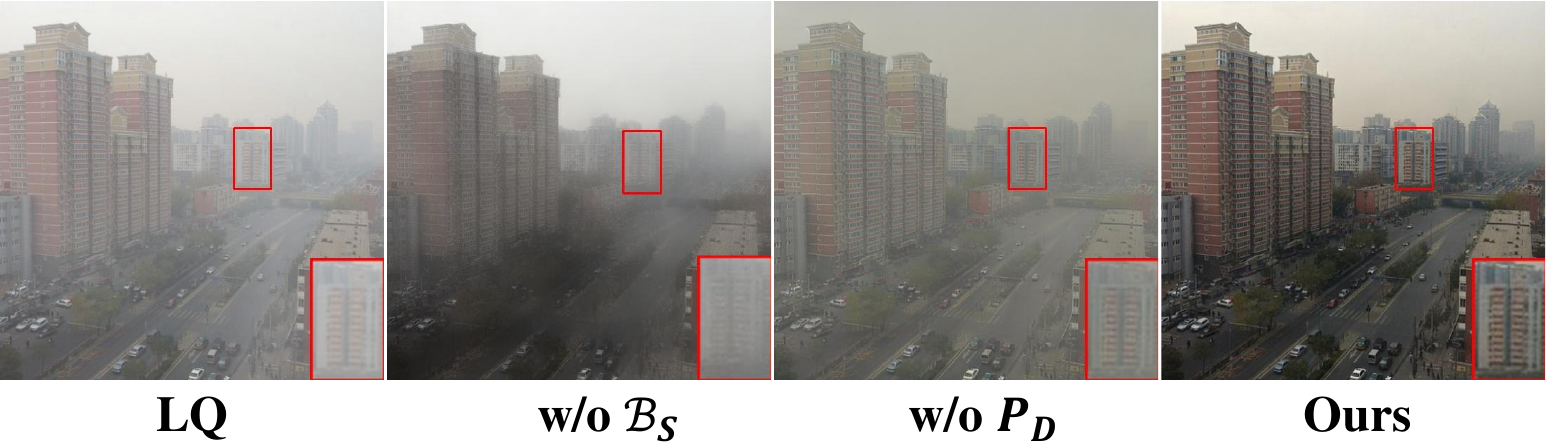}
\caption{Qualitative comparison of ablation study on visual prompts. The semantic branch can help to generate clear details and the degradation embedding plays a crucial role in degradation removal. \textbf{Please zoom in for the best view.}}
\vspace{-7mm}
\label{fig: ab_promptprocessor}
\end{figure}

\begin{figure*}[ht] 
\centering
\includegraphics[width=0.98\textwidth]{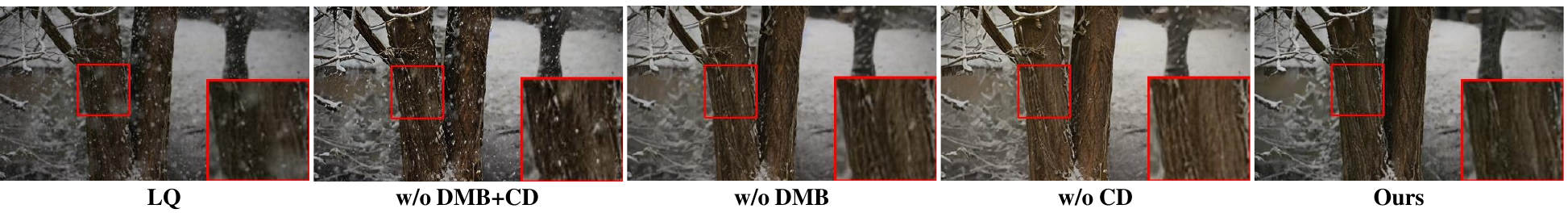}
\vspace{-2mm}
\caption{Qualitative comparison for Diff-Restorer with different control module. The designed Image-guided Control Module plays a crucial role in enabling the model to remove degradation and generate high-quality images that are more faithful to the original ones. \textbf{Please zoom in for the best view.}}
\vspace{-3mm}
\label{fig: ab_controlmodule}
\end{figure*}

\begin{figure*}[ht] 
\centering
\includegraphics[width=0.98\textwidth]{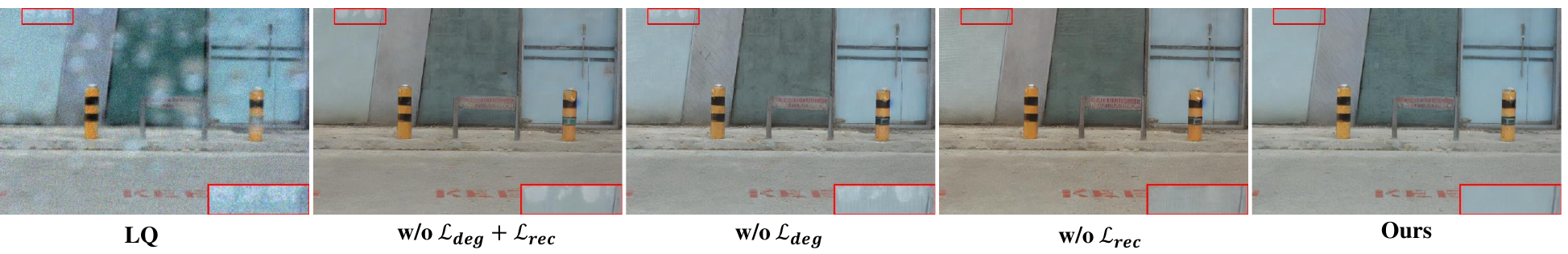}
\vspace{-2mm}
\caption{Qualitative comparison for Diff-Restorer training with different loss functions. The configuration of our loss functions enables the model to possess better color preservation capability and enhanced degradation removal ability. \textbf{Please zoom in for the best view.}}
\vspace{-3mm}
\label{fig: ab_loss}
\end{figure*}

\begin{table}[th]
\centering
\caption{Ablation studies on the Control Module. The metrics are reported on the average of the restoration tasks with ``low-light", ``haze", ``realsnow", ``snow+noise" and ``raindrop+blur+noise" degradations.}
\vspace{-2mm}
\label{tab:ab_controlmodule}
\resizebox{0.40\textwidth}{!}{
\begin{tabular}{lcccc} 
\toprule
Method & FID$\downarrow$ & LPIPS$\downarrow$ & DISTS$\downarrow$ & NIQE$\downarrow$ \\ 
\midrule
w/o DMB $+$ CD & 61.30 & 0.2385 & 0.1507  & 3.89  \\ 
w/o DMB & 49.64 & 0.2019 &  0.1295 & 3.76 \\ 
w/o CD & 52.07 & 0.2202 & 0.1413 & 3.86 \\ 
Ours$_{S}$ & 49.28 & 0.1989 & 0.1308 & 3.71 \\
Ours & 34.23 & 0.1331 & 0.0999 & 3.57 \\ 
\bottomrule
\end{tabular}}
\vspace{-3mm}
\end{table}
\begin{table}[th]
\centering
\caption{Ablation studies on the loss settings. The metrics are reported on the average of the restoration tasks with ``low-light", ``haze", ``realsnow", ``snow+noise" and ``raindrop+blur+noise" degradations.}
\vspace{-2mm}
\label{tab:ab_loss}
\resizebox{0.40\textwidth}{!}{
\begin{tabular}{lcccc} 
\toprule
Method & FID$\downarrow$ & LPIPS$\downarrow$ & DISTS$\downarrow$ & NIQE$\downarrow$ \\ 
\midrule
w/o $\mathcal{L}_{deg} + \mathcal{L}_{rec}$ & 52.68 & 0.2261 & 0.1437 & 4.05 \\  
w/o $\mathcal{L}_{deg}$ & 53.78 & 0.2218 &  0.1442 & 4.05 \\ 
w/o $\mathcal{L}_{rec}$ & 52.07 & 0.2202 &  0.1413 & 3.86 \\ 
Ours$_{S}$ & 49.28 & 0.1989 & 0.1308 & 3.71 \\
Ours & 34.23 & 0.1331 & 0.0999 & 3.57 \\ 
\bottomrule
\end{tabular}}
\vspace{-3mm}
\end{table}
\begin{table}[ht]
\centering
\caption{Ablation studies on the different VAE decoders. The metrics are reported on the average of the restoration tasks with ``low-light", ``haze", ``realsnow", ``snow+noise" and ``raindrop+blur+noise" degradations.}
\vspace{-2mm}
\label{tab:ab_decoder}
\resizebox{0.48\textwidth}{!}{
\begin{tabular}{lcccc}
\toprule
Method & FID$\downarrow$ & LPIPS$\downarrow$ & DISTS$\downarrow$ & NIQE$\downarrow$ \\ 
\midrule
SD decoder\cite{rombach2022ldm} & 49.28 & 0.1989 & 0.1308 & 3.71 \\ 
Consistency Decoder\cite{Betkerconsistency} & 49.61 & 0.2046 & 0.1320  & 3.50 \\
Ours decoder w/o DRB & 61.32  & 0.2291 & 0.1547 & 4.42 \\ 
Ours decoder & 34.23 & 0.1331 &  0.0999 & 3.57 \\ 
\bottomrule
\end{tabular}}
\vspace{-3mm}
\end{table}

\subsection{Ablation Study}
We conducted ablation studies on the proposed module to examine the effectiveness of our Diff-Restorer.

\textbf{Effectiveness of Visual Prompt Processor.} To ensure that we capture degradation-specific attributes, we first conducted an evaluation of the degradation embedding. We visualized the t-SNE plots of CLIP image embeddings and the degradation embeddings we extracted in Fig.\ref{fig: t-SNE}. It can be observed that compared to the original CLIP image embeddings, the degradation embeddings exhibit stronger task discriminability. This indicates that our method has effectively learned visual cues that enable the network to distinguish different types of degradations. This process does not require manual intervention, showcasing the adaptive capability of our model to discern and address degradation automatically. We also conducted experiments on the effectiveness of the semantic branch in extracting degradation-agnostic information and the validity of the degradation embedding. The results are shown in Table \ref{tab:ab_promptprocessor} and Fig. \ref{fig: ab_promptprocessor}. When we removed the semantic branch $\mathcal{B}_s$ and directly fed the CLIP image embedding $P_{CLIP}$ into the diffusion process, we observed that the generated images were very blurry and failed to effectively remove the degradation. This demonstrates that feeding $P_{CLIP}$ containing degradation into the diffusion process leads to undesirable results and can even conflict with the control module, resulting in worsened results. In contrast, our $\mathcal{B}_s$, under the implicit supervision of the overall diffusion process optimization objective, effectively extracts degradation-agnostic semantic information and help the diffusion generate more details. To validate the effectiveness of the degradation embedding $P_D$ in control module, we set it null. Due to the crucial role of $P_D$ in the Image-guided Control Module, when $P_D$ is null, degradation removal cannot be achieved.

\textbf{Effectiveness of Image-guided Control Module.}
We conducted experiments to evaluate the effectiveness of the Image-guided Control Module. The core components of our control module include Degradation Modulation Block (DMB) and Control Decoder (CD). We separately validated the effectiveness of these core components, and the experimental results are shown in Table \ref{tab:ab_controlmodule} and Fig. \ref{fig: ab_controlmodule}. It can be observed that removing the DMB and CD modules leads to a significant decrease in metrics such as FID, LPIPS, DISTS, and NIQE. This greatly reduces the quality and faithfulness of the restored images, particularly when the Control Decoder is removed. This demonstrates that our DMB and CD modules can extract useful information from low-quality images to guide the diffusion model in generating high-quality images that are faithful to the original images. Subjectively, as shown in Fig.\ref{fig: ab_controlmodule}, the introduction of DMB helps the model adaptively remove degradation and improve its degradation removal capability (column 3). The inclusion of CD helps the model maintain faithfulness to the original image, particularly in terms of color fidelity (column 4). These results validate the effectiveness of the Image-guided Control Module.

\textbf{Effectiveness of Loss Constraints.}
In the diffusion process, we introduced another two losses $\mathcal{L}_{deg}$ and $\mathcal{L}_{rec}$ to constrain the learning of our diffusion process. We conducted ablation experiments to evaluate the impact of these two losses, and the results are shown in Table \ref{tab:ab_loss} and Fig. \ref{fig: ab_loss}. It can be observed that removing the $\mathcal{L}_{deg}$ and $\mathcal{L}_{rec}$ leads to the decrease of FID, LPIPS, DISTS, and NIQE metrics and significantly reduces the faithfulness of the generated images to the original images and the perceptual quality of the generation results. With the constraints of these two losses, our method can better incorporate effective information extracted from the original low-quality images into the diffusion model, ensuring the generation of high-quality images that are faithful to the originals. As shown in Fig. \ref{fig: ab_loss}, removing $\mathcal{L}_{deg}$ loss leads to a decrease in the degradation removal capability, as seen in the raindrop example (column 3). On the other hand, removing the $\mathcal{L}_{rec}$ loss introduces color-shifting issues. Removing both losses not only fails to completely remove the degradation but also alters the image colors. Our approach leverages the combination of these two losses to address these challenges and achieves a good balance between realism and fidelity, resulting in impressive restoration performance.

\textbf{Effectiveness of Degradation-aware Decoder.}
Finally, we conducted experiments to evaluate the effectiveness of the Degradation-aware Decoder. We compared it with the SD Decoder\cite{rombach2022ldm}, Consistency Decoder\cite{Betkerconsistency}, and the Degradation-aware Decoder without the DRB component. Consistency Decoder is trained by OpenAI to be used to improve decoding for Stable Diffusion VAEs. The experimental results are shown in Table \ref{tab:ab_decoder} and Fig. \ref{fig: ab_decoder}. Quantitatively, our method achieves significant improvements in evaluation metrics such as FID, LPIPS, DISTS, and NIQE, compared to other methods. Additionally, when the DRB module is removed, the model cannot extract effective cues from low-quality images, and directly finetuning on the training data may reduce the model's generalization ability, leading to a decrease in evaluation metrics. Furthermore, as shown in Fig. \ref{fig: ab_decoder}, our Decoder demonstrates good correction capabilities, effectively compensating for the shortcomings of the original SD Decoder, especially in areas such as text (row 2)  and (row 3), significantly improving the quality of image reconstruction.

\section{Conclusion and Limitations}
In this work, we propose Diff-Restorer, a carefully designed architecture that leverages the remarkable generative priors encapsulated in pre-trained Stable Diffusion models to address universal image restoration problems. Our Diff-Restorer utilizes the image understanding capabilities of visual language models to adaptively mine visual prompts that drive the diffusion model for controllable generation. Additionally, to ensure fidelity to the original image, we design a Image-guided Control Module for spatial structure and color control. Furthermore, we introduce a Degradation-aware Decoder to address geometric distortions when mapping back from the latent space to the pixel domain in Stable Diffusion. Overall, our Diff-Restorer achieves adaptive handling of various degradation tasks and delivers high-quality restoration results. It demonstrates convincing performance in single degradation restoration tasks, real-world degradation restoration tasks, and mixed degradation restoration tasks both qualitatively and quantitatively.

One major limitation of our method is the longer inference times due to the use of diffusion model. To address this, future work can explore distillation strategies to improve the inference speed and enhance the usability of the model. Additionally, since our method leverages the diffusion model to enhance the perceptual quality of the images, it performs better on perceptual metrics but may exhibit average performance on metrics such as PSNR. Therefore, it is important to explore evaluation metrics that better represent image perceptual quality. Furthermore, our method's performance on certain image restoration tasks may be average due to the scale of the training dataset. Therefore, constructing larger and higher-quality datasets is an important direction to explore in the future.

\bibliographystyle{IEEEtran}
\bibliography{reference.bib}

\end{document}